\documentclass{article}

\pdfoutput=1

\usepackage{microtype}
\usepackage{graphicx}

\usepackage{url}
\usepackage{amsmath}
\usepackage{amsthm}
\usepackage{amssymb}
\usepackage{amsfonts}
\usepackage{subcaption}
\usepackage{bm}
\usepackage{hyperref}

\DeclareMathOperator*{\argmax}{arg\,max}
\newcommand{\diag}{\operatorname{diag}}
\usepackage{booktabs} % for professional tables

\usepackage[accepted]{icml2018}

\icmltitlerunning{Iterative Amortized Inference}

\begin{document}

\twocolumn[
\icmltitle{Iterative Amortized Inference}



\icmlsetsymbol{equal}{*}

\begin{icmlauthorlist}
\icmlauthor{Joseph Marino}{caltech}
\icmlauthor{Yisong Yue}{caltech}
\icmlauthor{Stephan Mandt}{disney}
\end{icmlauthorlist}

\icmlaffiliation{caltech}{California Institute of Technology (Caltech), Pasadena, CA, USA}
\icmlaffiliation{disney}{Disney Research, Los Angeles, CA, USA}

\icmlcorrespondingauthor{Joseph Marino}{jmarino@caltech.edu}

\icmlkeywords{Machine Learning, ICML}

\vskip 0.3in
]



\printAffiliationsAndNotice{}  % leave blank if no need to mention equal contribution

\begin{abstract}
Inference models are a key component in scaling variational inference to deep latent variable models, most notably as encoder networks in variational auto-encoders (VAEs). By replacing conventional optimization-based inference with a learned model, inference is amortized over data examples and therefore more computationally efficient. However, standard inference models are restricted to direct mappings from data to approximate posterior estimates. The failure of these models to reach fully optimized approximate posterior estimates results in an \textit{amortization gap}. We aim toward closing this gap by proposing \textit{iterative inference models}, which learn to perform inference optimization through repeatedly encoding gradients. Our approach generalizes standard inference models in VAEs and provides insight into several empirical findings, including top-down inference techniques. We demonstrate the inference optimization capabilities of iterative inference models and show that they outperform standard inference models on several benchmark data sets of images and text.
\end{abstract}

\section{Introduction}
\label{sec: introduction}

Variational inference \cite{jordan1998introduction} has been essential in learning deep directed latent variable models on high-dimensional data, enabling extraction of complex, non-linear relationships, such as object identities \cite{higgins2016beta} and dynamics \cite{xue2016visual, karl2016deep} directly from observations. Variational inference reformulates inference as optimization \cite{ neal1998view, hoffman2013stochastic}. However, the current trend has moved toward employing \textit{inference models} \cite{dayan1995helmholtz, gregor2013deep, kingma2014stochastic, rezende2014stochastic}, mappings from data to approximate posterior estimates that are amortized across examples. Intuitively, the inference model encodes observations into latent representations, and the generative model decodes these representations into reconstructions. Yet, this approach has notable limitations. For instance, in models with empirical priors, such as hierarchical latent variable models, ``bottom-up" data-encoding inference models cannot account for ``top-down" priors (Section \ref{sec: hierarchical models}). This has prompted the use of top-down inference techniques \cite{sonderby2016ladder}, which currently lack a rigorous theoretical justification. More generally, the inability of inference models to reach fully optimized approximate posterior estimates results in decreased modeling performance, referred to as an \textit{amortization gap} \cite{krishnan2017challenges, cremerinference}.

To combat this problem, our work offers a departure from previous approaches by re-examining inference from an optimization perspective. We utilize approximate posterior gradients to perform inference optimization. Yet, we improve computational efficiency over conventional optimizers by encoding these gradients with an inference model that learns how to iteratively update approximate posterior estimates. The resulting \textit{iterative inference models} resemble learning to learn \cite{andrychowicz2016learning} applied to variational inference optimization. However, we refine and extend this method along several novel directions. Namely, (1) we show that learned optimization models can be applied to inference optimization of latent variables; (2) we show that non-recurrent optimization models work well in practice, breaking assumptions about the necessity of non-local curvature for outperforming conventional optimizers \cite{andrychowicz2016learning, putzky2017recurrent}; and (3) we provide a new form of optimization model that encodes errors rather than gradients to approximate higher order derivatives, empirically resulting in faster convergence.

Our main contributions are summarized as follows:
\begin{enumerate}
    \item we introduce a family of iterative inference models, which generalize standard inference models,
    \item we provide the first theoretical justification for top-down inference techniques,
    \item we empirically evaluate iterative inference models, demonstrating that they outperform standard inference models on several data sets of images and text.
\end{enumerate}

\section{Background}
\label{sec: background}

\subsection{Latent Variable Models \& Variational Inference}
\label{sec: latent variable models and variational inference}

Latent variable models are generative probabilistic models that use local (per data example) latent variables, $\mathbf{z}$, to model observations, $\mathbf{x}$, using global (across data examples) parameters, $\theta$. A model is defined by the joint distribution $p_\theta (\mathbf{x}, \mathbf{z}) = p_\theta (\mathbf{x} | \mathbf{z}) p_\theta (\mathbf{z})$, composed of the conditional likelihood and the prior. Learning the model parameters and inferring the posterior, $p(\mathbf{z} | \mathbf{x})$, are intractable for all but the simplest models, as they require evaluating the marginal likelihood, $p_\theta (\mathbf{x}) = \int p_\theta (\mathbf{x}, \mathbf{z}) d\mathbf{z}$, which involves integrating the model over $\mathbf{z}$. For this reason, we often turn to approximate inference methods.

Variational inference reformulates this intractable integration as an optimization problem by introducing an approximate posterior\footnote{We use $q(\mathbf{z}|\mathbf{x})$ to denote that the approximate posterior is conditioned on a data example (i.e. local), however this does not necessarily imply a direct functional mapping.}, $q(\mathbf{z} | \mathbf{x})$, typically chosen from some tractable family of distributions, and minimizing the KL-divergence from the posterior, $D_{KL} (q(\mathbf{z} | \mathbf{x}) || p(\mathbf{z} | \mathbf{x}))$. This quantity cannot be minimized directly, as it contains the posterior. Instead, KL-divergence can be decomposed into
\begin{equation}
    D_{KL} (q(\mathbf{z} | \mathbf{x}) || p(\mathbf{z} | \mathbf{x})) = \log p_\theta (\mathbf{x}) -\mathcal{L},
    \label{eq: KL decomposition}
\end{equation}
where $\mathcal{L}$ is the evidence lower bound (ELBO), which is defined as:
\begin{align}
    \mathcal{L} & \equiv \mathbb{E}_{\mathbf{z} \sim q(\mathbf{z} | \mathbf{x})} \left[ \log p_\theta (\mathbf{x}, \mathbf{z}) - \log q(\mathbf{z} | \mathbf{x}) \right]
    \label{eq: ELBO definition} \\
    & = \mathbb{E}_{\mathbf{z} \sim q(\mathbf{z} | \mathbf{x})} \left[ \log p_\theta (\mathbf{x} | \mathbf{z}) \right] - D_{KL} (q(\mathbf{z} | \mathbf{x}) || p_\theta (\mathbf{z})).
    \label{eq: ELBO definition 2}
\end{align}
The first term in eq. \ref{eq: ELBO definition 2} expresses how well the output reconstructs the data example. The second term quantifies the dissimilarity between the approximate posterior and the prior. Because $\log p_\theta (\mathbf{x})$ is not a function of $q(\mathbf{z} | \mathbf{x})$, we can minimize $D_{KL} (q(\mathbf{z} | \mathbf{x}) || p(\mathbf{z} | \mathbf{x}))$ in eq. \ref{eq: KL decomposition} by maximizing $\mathcal{L}$ w.r.t. $q(\mathbf{z} | \mathbf{x})$, thereby performing approximate \textit{inference}. Likewise, because $D_{KL} (q(\mathbf{z} | \mathbf{x}) || p(\mathbf{z} | \mathbf{x}))$ is non-negative, $\mathcal{L}$ is a lower bound on $\log p_\theta (\mathbf{x})$. Therefore, once we have inferred an optimal $q(\mathbf{z} | \mathbf{x})$, \textit{learning} corresponds to maximizing $\mathcal{L}$ w.r.t. $\theta$.

\subsection{Variational Expectation Maximization (EM) via Gradient Ascent}

The optimization procedures for variational inference and learning are respectively the expectation and maximization steps of the variational EM algorithm \cite{dempster1977maximum, neal1998view}, which alternate until convergence. This is typically performed in the batched setting of stochastic variational inference \cite{hoffman2013stochastic}. When $q(\mathbf{z} | \mathbf{x})$ takes a parametric form, the expectation step for data example $\mathbf{x}^{(i)}$ involves finding a set of distribution parameters, $\bm{\lambda}^{(i)}$, that are optimal w.r.t. $\mathcal{L}$. With a factorized Gaussian density over continuous latent variables, i.e. $\bm{\lambda}^{(i)} = \{ \bm{\mu}^{(i)}_q , \bm{\sigma}_q^{2(i)} \}$ and $q(\mathbf{z}^{(i)} | \mathbf{x}^{(i)}) = \mathcal{N} (\mathbf{z}^{(i)}; \bm{\mu}_q^{(i)}, \diag{\bm{\sigma}_q^{2(i)}})$, conventional optimization techniques repeatedly estimate the stochastic gradients $\nabla_{\bm{\lambda}} \mathcal{L}$ to optimize $\mathcal{L}$ w.r.t. ${\bm{\lambda}}^{(i)}$, e.g.:
\begin{equation}
    {\bm{\lambda}}^{(i)} \leftarrow {\bm{\lambda}}^{(i)} + \alpha \nabla_{\bm{\lambda}} \mathcal{L} (\mathbf{x}^{(i)}, {\bm{\lambda}}^{(i)}; \theta),
    \label{eq: svi update}
\end{equation}
where $\alpha$ is the step size. This procedure, which is repeated for each example, is computationally expensive and requires setting step-size hyper-parameters.

\subsection{Amortized Inference Models}
\label{sec: amortized inference}

Due to the aforementioned issues, gradient updates of approximate posterior parameters are rarely performed in practice. Rather, inference models are often used to map observations to approximate posterior estimates. Optimization of each data example's approximate posterior parameters, $\bm{\lambda}^{(i)}$, is replaced with the optimization of a shared, i.e. amortized \cite{gershman2014amortized}, set of parameters, $\phi$, contained within an inference model, $f$, of the form:
\begin{equation}
    \bm{\lambda}^{(i)} \leftarrow f(\mathbf{x}^{(i)}; \phi).
    \label{eq: inference model form}
\end{equation}
While inference models have a long history, e.g. \cite{dayan1995helmholtz}, the most notable recent example is the variational auto-encoder (VAE) \cite{kingma2014stochastic, rezende2014stochastic}, which employs the reparameterization trick to propagate stochastic gradients from the generative model to the inference model, both of which are parameterized by neural networks. We refer to inference models of this form as \textit{standard inference models}. As discussed in Section \ref{sec: iterative amortized inference}, the aim of this paper is to move beyond the direct encoder paradigm of standard inference models to develop improved techniques for performing inference.

\section{Iterative Amortized Inference}
\label{sec: iterative amortized inference}

In Section \ref{iterative inference models}, we introduce our contribution, iterative inference models. However, we first motivate our approach in Section \ref{sec: inference models perform approximate optimization} by discussing the limitations of standard inference models. We then draw inspiration from other techniques for learning to optimize (Section \ref{sec: learning to iteratively optimize}).

\begin{figure*}[t!]
\centering
\includegraphics[width=0.83\textwidth]{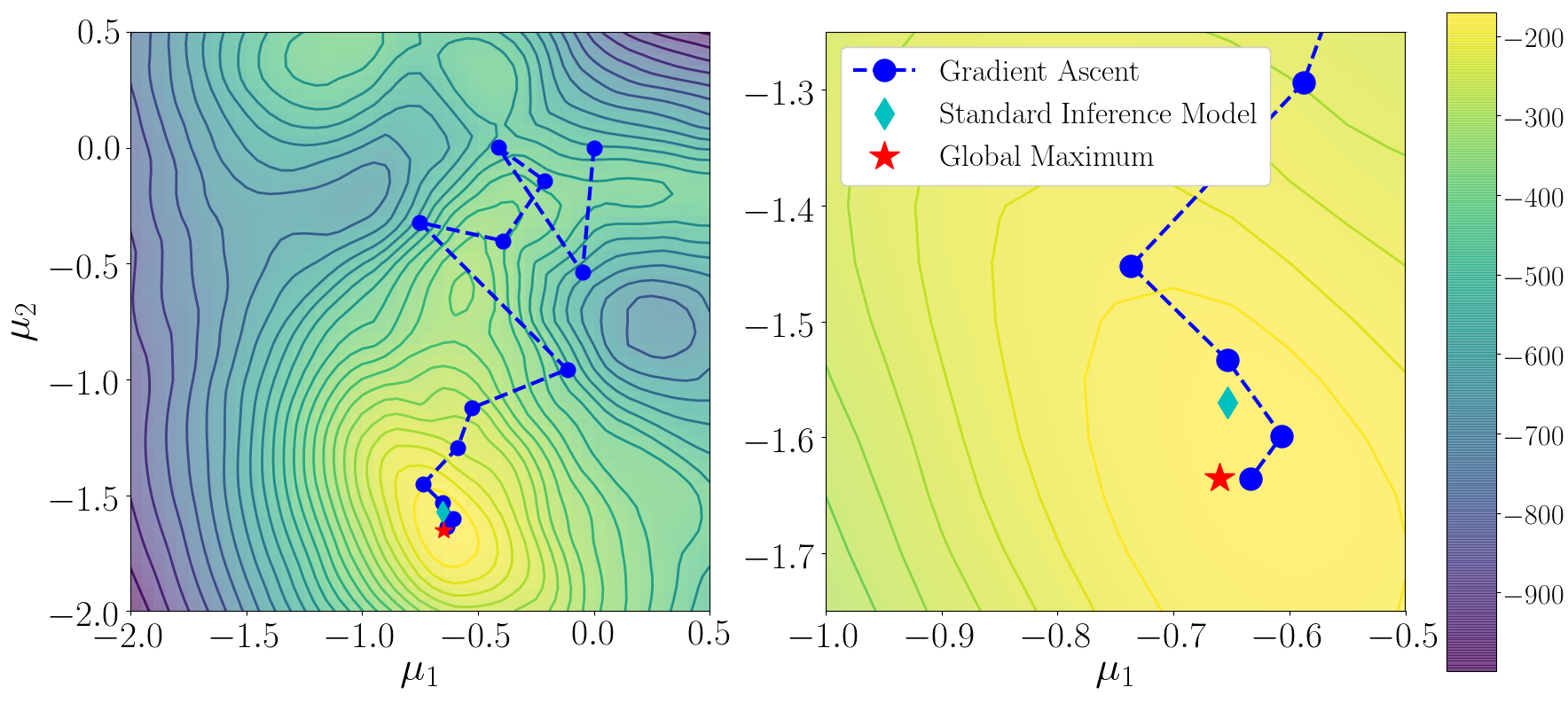}
\caption{\textbf{Visualizing the amortization gap.} Optimization surface of $\mathcal{L}$ (in \textit{nats}) for a 2-D latent Gaussian model and an MNIST data example. Shown on the plots are the optimal estimate (MAP), the output of a standard inference model, and an optimization trajectory of gradient ascent. The plot on the right shows an enlarged view near the optimum. Conventional optimization outperforms the standard inference model, exhibiting an amortization gap. With additional latent dimensions or more complex data, this gap could become larger.}
\label{fig: ELBO optimization surface}
\end{figure*}

\subsection{Standard Inference Models \& Amortization Gaps}
\label{sec: inference models perform approximate optimization}

As described in Section \ref{sec: latent variable models and variational inference}, variational inference reformulates inference as the maximization of $\mathcal{L}$ w.r.t. $q(\mathbf{z} | \mathbf{x})$, constituting the expectation step of the variational EM algorithm. In general, this is a difficult non-convex optimization problem, typically requiring a lengthy iterative estimation procedure. Yet, standard inference models attempt to perform this optimization through a direct, discriminative mapping from data observations to approximate posterior parameters. Of course, generative models can adapt to accommodate sub-optimal approximate posteriors. Nevertheless, the possible limitations of a direct inference mapping applied to this difficult optimization procedure may result in a decrease in overall modeling performance.

We demonstrate this concept in Figure \ref{fig: ELBO optimization surface} by visualizing the optimization surface of $\mathcal{L}$ defined by a 2-D latent Gaussian model and a particular binarized MNIST \cite{lecun1998gradient} data example. To visualize the approximate posterior, we use a point estimate, $q(\mathbf{z}|\mathbf{x}) = \delta (\bm{\mu}_q)$, where $\bm{\mu}_q = (\mu_1, \mu_2)$ is the estimate and $\delta$ is the Dirac delta function. See Appendix C.1 for details. Shown on the plot are the optimal (maximum a posteriori or MAP) estimate, the estimate from a standard inference model, and an optimization trajectory of gradient ascent. The inference model is unable to achieve the optimum, but manages to output a reasonable estimate in one pass. Gradient ascent requires many iterations and is sensitive to step-size, but through the iterative estimation procedure, ultimately arrives at a better final estimate. The inability of inference models to reach optimal approximate posterior estimates, as typically compared with gradient-based methods, creates an amortization gap \cite{krishnan2017challenges, cremerinference}, which impairs modeling performance. Additional latent dimensions and more complex data could further exacerbate this problem. 

\subsection{Learning to Iteratively Optimize}
\label{sec: learning to iteratively optimize}

While offering significant benefits in computational efficiency, standard inference models can suffer from sizable amortization gaps \cite{krishnan2017challenges}. Parameterizing inference models as direct, static mappings from $\mathbf{x}$ to $q(\mathbf{z} | \mathbf{x})$ may be overly restrictive, widening this gap. To improve upon this direct encoding paradigm, we pose the following question: \textit{can we retain the computational efficiency of inference models while incorporating more powerful iterative estimation capabilities?} Our proposed solution is a new class of inference models, capable of learning how to update approximate posterior estimates by encoding gradients or errors. Due to the iterative nature of these models, we refer to them as \textit{iterative inference models}. Through an analysis with latent Gaussian models, we show that iterative inference models generalize standard inference models (Section \ref{sec: relation to standard inference models}) and offer theoretical justification for top-down inference in hierarchical models (Section \ref{sec: hierarchical models}).

Our approach relates to learning to learn \cite{andrychowicz2016learning}, where an \textit{optimizer} model learns to optimize the parameters of an \textit{optimizee} model. The optimizer receives the optimizee's parameter gradients and outputs updates to these parameters to improve the optimizee's loss. The optimizer itself can be learned due to the differentiable computation graph. Such models can adaptively adjust step sizes, potentially outperforming conventional optimizers. For inference optimization, previous works have combined standard inference models with gradient updates \cite{hjelm2016iterative, krishnan2017challenges, kim2018semi}, however, these works do not \textit{learn} to iteratively optimize. \cite{putzky2017recurrent} use recurrent inference models for MAP estimation of denoised images in linear models. We propose a unified method for learning to perform variational inference optimization, generally applicable to probabilistic latent variable models. Our work extends techniques for learning to optimize along several novel directions, discussed in Section \ref{iterative inference in latent gaussian models}.

\subsection{Iterative Inference Models}
\label{iterative inference models}

We denote an iterative inference model as $f$ with parameters $\phi$. With $\mathcal{L}^{(i)}_t \equiv \mathcal{L}(\mathbf{x}^{(i)}, \bm{\lambda}^{(i)}_{t}; \theta)$ as the ELBO for data example $\mathbf{x}^{(i)}$ at inference iteration $t$, the model uses the approximate posterior gradients, denoted $\nabla_{\bm{\lambda}} \mathcal{L}^{(i)}_t$, to output updated estimates of $\bm{\lambda}^{(i)}$:
\begin{equation}
    \bm{\lambda}^{(i)}_{t+1} \leftarrow f_t (\nabla_{\bm{\lambda}} \mathcal{L}^{(i)}_t, \bm{\lambda}^{(i)}_{t}; \phi),
    \label{eq: iterative inference form}
\end{equation}
where $\bm{\lambda}^{(i)}_t$ is the estimate of $\bm{\lambda}^{(i)}$ at inference iteration $t$. Eq. \ref{eq: iterative inference form} is in a general form and contains, as special cases, the linear update in eq. \ref{eq: svi update}, as well as the residual, non-linear update used in \cite{andrychowicz2016learning}. 
Figure \ref{fig: computation graph} displays a computation graph of the inference procedure, and Algorithm 1 in Appendix B describes the procedure in detail. As with standard inference models, the parameters of an iterative inference model can be updated using estimates of $\nabla_\phi \mathcal{L}$, obtained through the reparameterization trick \cite{kingma2014stochastic, rezende2014stochastic} or through score function methods \cite{gregor2013deep, ranganath2014black}. Model parameter updating is performed using stochastic gradient techniques with $\nabla_\theta \mathcal{L}$ and $\nabla_\phi \mathcal{L}$.

\begin{figure}[t!]
    \centering
    \includegraphics[width=0.37\textwidth]{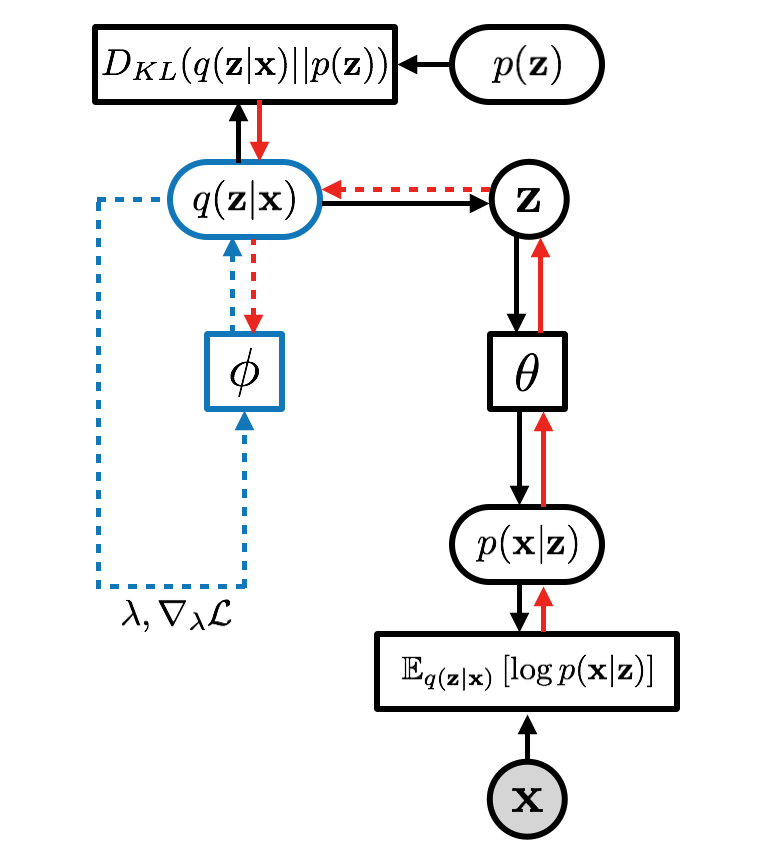}
    \caption{\textbf{Computation graph} for a single-level latent variable model with an iterative inference model. Black components evaluate the ELBO. Blue components are used during variational inference. Red corresponds to gradients. Solid arrows denote deterministic values. Dashed arrows denote stochastic values. During inference, $\bm{\lambda}$, the distribution parameters of $q(\mathbf{z} | \mathbf{x})$, are first initialized. $\mathbf{z}$ is sampled from $q(\mathbf{z} | \mathbf{x})$ to evaluate the ELBO. Stochastic gradients are then backpropagated to $\bm{\lambda}$. The iterative inference model uses these gradients to update the current estimate of $\bm{\lambda}$. The process is repeated iteratively. The inference model parameters, $\phi$, are trained through accumulated estimates of $\nabla_\phi \mathcal{L}$.}
    \label{fig: computation graph}
\end{figure}

\section{Iterative Inference in Latent Gaussian Models}
\label{iterative inference in latent gaussian models}

We now describe an instantiation of iterative inference models for (single-level) latent Gaussian models, which have a Gaussian prior density over latent variables: $p(\mathbf{z}) = \mathcal{N} (\mathbf{z}; \bm{\mu}_p, \diag{\bm{\sigma}^2_p})$. Although the prior is typically a standard Normal density, we use this prior form for generality. Latent Gaussian models are often used in VAEs and are a common choice for continuous-valued latent variables. While the approximate posterior can be any probability density, it is typically also chosen as Gaussian: $q(\mathbf{z} | \mathbf{x}) = \mathcal{N} (\mathbf{z}; \bm{\mu}_q, \diag{\bm{\sigma}^2_q})$. With this choice, $\bm{\lambda}^{(i)}$ corresponds to $\{ \bm{\mu}^{(i)}_q , \bm{\sigma}^{2 (i)}_q \}$ for example $\mathbf{x}^{(i)}$. Dropping the superscript $(i)$ to simplify notation, we can express eq. \ref{eq: iterative inference form} for this model as:
\begin{equation}
    \bm{\mu}_{q, t+1} = f^{\bm{\mu}_q}_t (\nabla_{\bm{\mu}_q} \mathcal{L}_t, \bm{\mu}_{q, t}; \phi),
    \label{eq: mean iterative update}
\end{equation}
\begin{equation}
    \bm{\sigma}^{2}_{q, t+1} = f^{\bm{\sigma}^2_q}_t (\nabla_{\bm{\sigma}^2_q} \mathcal{L}_t, \bm{\sigma}^{2}_{q, t}; \phi),
    \label{eq: variance iterative update}
\end{equation}
where $f^{\bm{\mu}_q}_t$ and $f^{\bm{\sigma}^2_q}_t$ are the iterative inference models for updating $\bm{\mu}_q$ and $\bm{\sigma}^{2}_q$ respectively. In practice, these models can be combined, with shared inputs and model parameters but separate outputs to update each term.

In Appendix A, we derive the stochastic gradients $\nabla_{\bm{\mu}_q} \mathcal{L}$ and $\nabla_{\bm{\sigma}^2_q} \mathcal{L}$ for the cases where $p_\theta (\mathbf{x} | \mathbf{z})$ takes a Gaussian and Bernoulli form, though \textit{any} output distribution can be used. Generally, these gradients are comprised of (1) errors, expressing the mismatch in distributions, and (2) Jacobian matrices, which invert the generative mappings. For instance, assuming a Gaussian output density, $p(\mathbf{x} | \mathbf{z}) = \mathcal{N} (\mathbf{x}; \bm{\mu}_\mathbf{x}, \diag{\bm{\sigma}^2_\mathbf{x}})$, the gradient for $\bm{\mu}_q$ is
\begin{equation}
    \nabla_{\bm{\mu}_q} \mathcal{L} = \mathbf{J}^\intercal \bm{\varepsilon}_{\mathbf{x}} - \bm{\varepsilon}_{\mathbf{z}},
    \label{eq: latent gaussian mean gradient}
\end{equation}
where the Jacobian ($\mathbf{J}$), bottom-up errors ($\bm{\varepsilon}_{\mathbf{x}}$), and top-down errors ($\bm{\varepsilon}_{\mathbf{z}}$) are defined as
\begin{equation}
    \mathbf{J} \equiv \mathbb{E}_{\mathbf{z} \sim q (\mathbf{z} | \mathbf{x})} \left[ {\frac{\partial \bm{\mu}_\mathbf{x} }{\partial \bm{\mu}_q}} \right],
    \label{eq: jacobian def}
\end{equation}
\begin{equation}
    \bm{\varepsilon}_{\mathbf{x}} \equiv \mathbb{E}_{\mathbf{z} \sim q (\mathbf{z} | \mathbf{x})} [(\mathbf{x} - \bm{\mu}_{ \mathbf{x}}) / \bm{\sigma}_\mathbf{x}^2 ],
    \label{eq: obs error def}
\end{equation}
\begin{equation}
    \bm{\varepsilon}_{\mathbf{z}} \equiv \mathbb{E}_{\mathbf{z} \sim q (\mathbf{z} | \mathbf{x})} [(\mathbf{z} - \bm{\mu}_p) / \bm{\sigma}^2_p ].
    \label{eq: latent error def}
\end{equation}
Here, we have assumed $\bm{\mu}_\mathbf{x}$ is a function of $\mathbf{z}$ and $\bm{\sigma}^2_\mathbf{x}$ is a global parameter. The gradient $\nabla_{\bm{\sigma}^2_q} \mathcal{L}$ is comprised of similar terms as well as an additional term penalizing approximate posterior entropy. Inspecting and understanding the composition of the gradients reveals the forces pushing the approximate posterior toward agreement with the data, through $\bm{\varepsilon}_{\mathbf{x}}$, and agreement with the prior, through $\bm{\varepsilon}_{\mathbf{z}}$. In other words, \textit{inference is as much a top-down process as it is a bottom-up process}, and the optimal combination of these terms is given by the approximate posterior gradients. As discussed in Section \ref{sec: hierarchical models}, standard inference models have traditionally been purely bottom-up, only encoding the data.

\subsection{Reinterpreting Top-Down Inference}
\label{sec: hierarchical models}

To increase the model capacity of latent variable models, it is common to add higher-level latent variables, thereby providing flexible \textit{empirical priors} on lower-level variables. Traditionally, corresponding standard inference models were parmeterized as purely bottom-up (e.g. Fig.\ 1 of \cite{rezende2014stochastic}). It was later found to be beneficial to incorporate top-down information from higher-level variables in the inference model, the given intuition being that ``\textit{a purely bottom-up inference process \dots does not correspond well with real perception}" \cite{sonderby2016ladder}, however, a rigorous justification of this technique was lacking.

Iterative inference models, or rather, the gradients that they encode, provide a theoretical explanation for this previously empirical heuristic. As seen in eq. \ref{eq: latent gaussian mean gradient}, the approximate posterior parameters are optimized to agree with the prior, while also fitting the conditional likelihood to the data. Analogous terms appear in the gradients for hierarchical models. For instance, in a chain-structured hierarchical model, the gradient of $\bm{\mu}_q^{\ell}$, the approximate posterior mean at layer $\ell$, is
\begin{equation}
    \nabla_{\bm{\mu}^\ell_q} \mathcal{L} = \mathbf{J}^{\ell \intercal} \bm{\varepsilon}^{\ell-1}_{\mathbf{z}} - \bm{\varepsilon}^{\ell}_{\mathbf{z}},
    \label{eq: hierarchical mean gradient}
\end{equation}
where $\mathbf{J}^{\ell}$ is the Jacobian of the generative mapping at layer $\ell$ and $\bm{\varepsilon}^{\ell}_{\mathbf{z}}$ is defined similarly to eq. \ref{eq: latent error def}. $\bm{\varepsilon}^{\ell}_{\mathbf{z}}$ depends on the top-down prior at layer $\ell$, which, unlike the single-level case, varies across data examples. Thus, a purely bottom-up inference procedure may struggle, as it must model both the bottom-up data dependence as well as the top-down prior. Top-down inference \cite{sonderby2016ladder} explicitly uses the prior to perform inference. Iterative inference models instead rely on approximate posterior gradients, which naturally account for both bottom-up and top-down influences.

\subsection{Approximating Approximate Posterior Derivatives}
\label{sec: approximate gradients}

In the formulation of iterative inference models given in eq. \ref{eq: iterative inference form}, inference optimization is restricted to first-order approximate posterior derivatives. Thus, it may require many inference iterations to reach reasonable approximate posterior estimates. Rather than calculate costly higher-order derivatives, we can take a different approach.

Approximate posterior derivatives (e.g. eq. \ref{eq: latent gaussian mean gradient} and higher-order derivatives) are essentially defined by the errors at the current estimate, as the other factors, such as the Jacobian matrices, are internal to the model. Thus, the errors provide more general information about the curvature beyond the gradient. As iterative inference models already learn to perform approximate posterior updates, it is natural to ask whether the errors provide a sufficient signal for faster inference optimization. In other words, we may be able to offload approximate posterior derivative calculation onto the inference model, yielding a model that requires fewer inference iterations while maintaining or possibly improving computational efficiency.

Comparing with eqs. \ref{eq: mean iterative update} and \ref{eq: variance iterative update}, the form of this new iterative inference model is
\begin{equation}
    \bm{\mu}_{q, t+1} = f^{\bm{\mu}_q}_t (\bm{\varepsilon}_{\mathbf{x}, t}, \bm{\varepsilon}_{\mathbf{z}, t}, \bm{\mu}_{q, t}; \phi),
    \label{eq: mean approximate iterative update}
\end{equation}
\begin{equation}
    \bm{\sigma}^{2}_{q, t+1} = f^{\bm{\sigma}^2_q}_t (\bm{\varepsilon}_{\mathbf{x}, t}, \bm{\varepsilon}_{\mathbf{z}, t}, \bm{\sigma}^{2}_{q, t}; \phi),
    \label{eq: variance approximate iterative update}
\end{equation}
where, again, these models can be shared, with separate outputs per parameter. In Section \ref{sec: samples and iterations}, we empirically find that models of this form converge to better solutions than gradient-encoding models when given fewer inference iterations. It is also worth noting that this error encoding scheme is similar to DRAW \cite{gregor2015draw}. However, in addition to architectural differences in the generative model, DRAW and later extensions do not include top-down errors \cite{gregor2016towards}, nor error precision-weighting.

\subsection{Generalizing Standard Inference Models}
\label{sec: relation to standard inference models}

Under certain assumptions on single-level latent Gaussian models, iterative inference models of the form in Section \ref{sec: approximate gradients} \textit{generalize} standard inference models. First, note that $\bm{\varepsilon}_{\mathbf{x}}$ (eq. \ref{eq: obs error def}) is a stochastic affine transformation of $\mathbf{x}$:
\begin{equation}
    \bm{\varepsilon}_{\mathbf{x}} = \mathbf{A} \mathbf{x} + \mathbf{b},
\end{equation}
where
\begin{equation}
    \mathbf{A} \equiv \mathbb{E}_{q (\mathbf{z} | \mathbf{x})} \left[ (\diag \bm{\sigma}_\mathbf{x}^2 )^{-1} \right],
\end{equation}
\begin{equation}
    \mathbf{b} \equiv - \mathbb{E}_{q (\mathbf{z} | \mathbf{x})} \left[ \frac{ \bm{\mu}_\mathbf{x}}{\bm{\sigma}_\mathbf{x}^2} \right].
\end{equation}
Reasonably assuming that the initial approximate posterior and prior are both constant, then in expectation, $\mathbf{A}$, $\mathbf{b}$, and $\bm{\varepsilon}_{\mathbf{z}}$ are constant across all data examples at the first inference iteration. Using proper weight initialization and input normalization, it is equivalent to input $\mathbf{x}$ or an affine transformation of $\mathbf{x}$ into a fully-connected neural network. Therefore, \textit{standard inference models are equivalent to the special case of a one-step iterative inference model}. Thus, we can interpret standard inference models as learning a map of local curvature around a fixed approximate posterior estimate. Iterative inference models, on the other hand, learn to traverse the optimization landscape more generally.

\begin{figure*}[t!]
\centering
\includegraphics[width=0.95\textwidth]{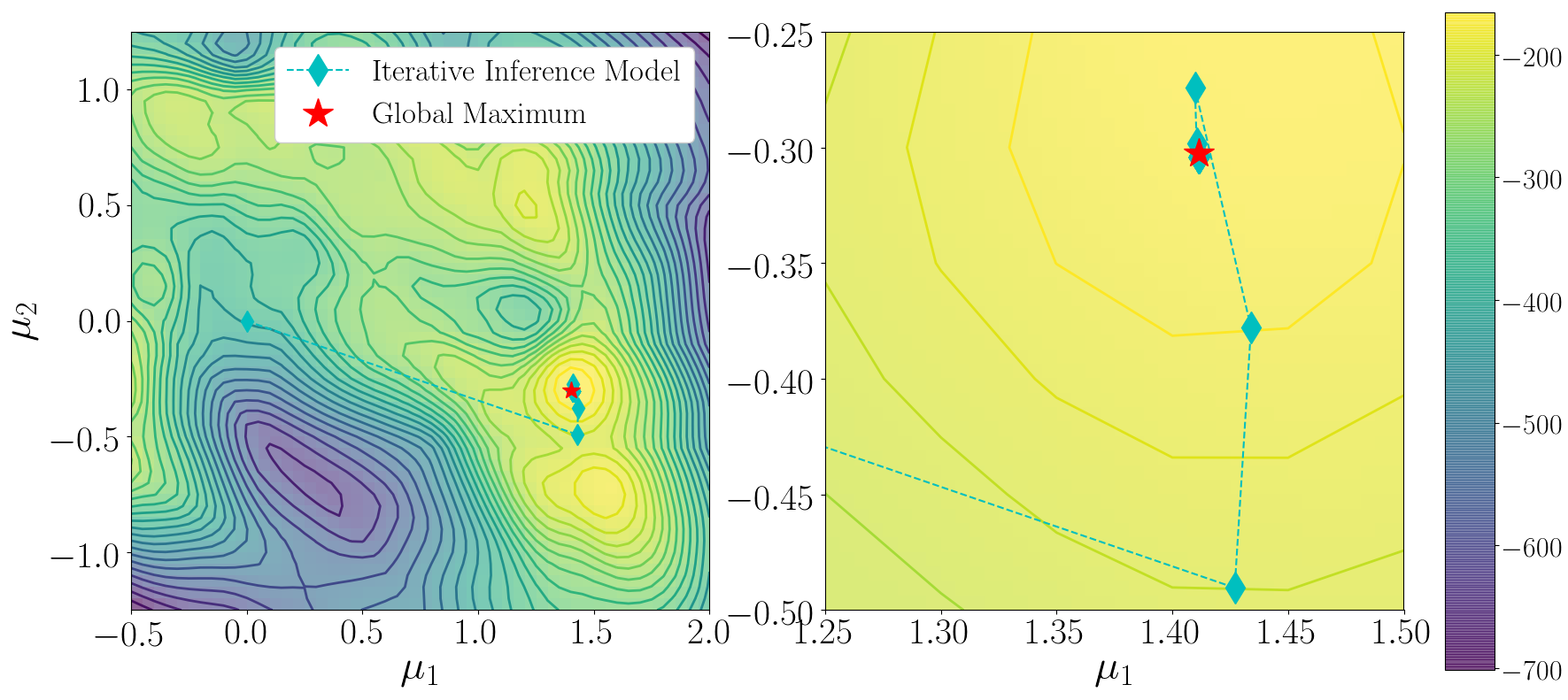}
\caption{\textbf{Direct visualization of iterative amortized inference optimization.} Optimization trajectory on $\mathcal{L}$ (in \textit{nats}) for an iterative inference model with a 2D latent Gaussian model for a particular MNIST example. The iterative inference model adaptively adjusts inference update step sizes to iteratively refine the approximate posterior estimate.}
\label{fig: iterative inference 2D optimization}
\end{figure*}

\section{Experiments}
\label{sec: experiments}

Using latent Gaussian models, we performed an empirical evaluation of iterative inference models on both image and text data. For images, we used \textbf{MNIST} \cite{lecun1998gradient}, \textbf{Omniglot} \cite{lake2013one}, \textbf{Street View House Numbers (SVHN)} \cite{netzer2011reading}, and \textbf{CIFAR-10} \cite{krizhevsky2009learning}. MNIST and Omniglot were dynamically binarized and modeled with Bernoulli output distributions, and SVHN and CIFAR-10 were modeled with Gaussian output densities, using the procedure from \cite{gregor2016towards}. For text, we used \textbf{RCV1} \cite{lewis2004rcv1}, with word count data modeled with a multinomial output.

Details on implementing iterative inference models are found in Appendix B. The primary difficulty of training iterative inference models comes from shifting gradient and error distributions during the course of inference and learning. In some cases, we found it necessary to normalize these inputs using layer normalization \cite{ba2016layer}. We also found it beneficial, though never necessary, to additionally encode the data itself, particularly when given few inference iterations (see Figure \ref{fig: additional iterations}). For comparison, all experiments use feedforward networks, though we observed similar results with recurrent inference models. Reported values of $\mathcal{L}$ were estimated using 1 sample, and reported values of $\log p (\mathbf{x})$ and perplexity (Tables \ref{tab: quant results} \& \ref{tab: text results}) were estimated using 5,000 importance weighted samples. Additional experiment details, including model architectures, can be found in Appendix C. Accompanying code can be found on GitHub at \href{https://github.com/joelouismarino/iterative_inference}{\texttt{joelouismarino/iterative\_inference}}.

Section \ref{sec: approximate optimization results} demonstrates the optimization capabilities of iterative inference models. Section \ref{sec: samples and iterations} explores two methods by which to further improve the modeling performance of these models. Section \ref{sec: comparison with standard inference models} provides a quantitative comparison between standard and iterative inference models.

\subsection{Approximate Inference Optimization}
\label{sec: approximate optimization results}

We begin with a series of experiments that demonstrate the inference optimization capabilities of iterative inference models. These experiments confirm that iterative inference models indeed learn to perform inference optimization through an adaptive iterative estimation procedure. These results highlight the qualitative differences between this inference optimization procedure and that of standard inference models. That is, iterative inference models are able to effectively utilize multiple inference iterations rather than collapsing to static, one-step encoders.

\textbf{Direct Visualization}
As in Section \ref{sec: inference models perform approximate optimization}, we directly visualize iterative inference optimization in a 2-D latent Gaussian model trained on MNIST with a point estimate approximate posterior. Model architectures are identical to those used in Section \ref{sec: inference models perform approximate optimization}, with additional details found in Appendix C.1. Shown in Figure \ref{fig: iterative inference 2D optimization} is a 16-step inference optimization trajectory taken by the iterative inference model for a particular example. The model adaptively adjusts inference update step sizes to navigate the optimization surface, quickly arriving and remaining at a near-optimal estimate.

\begin{figure}[t!]
    \centering
    \includegraphics[width=\linewidth]{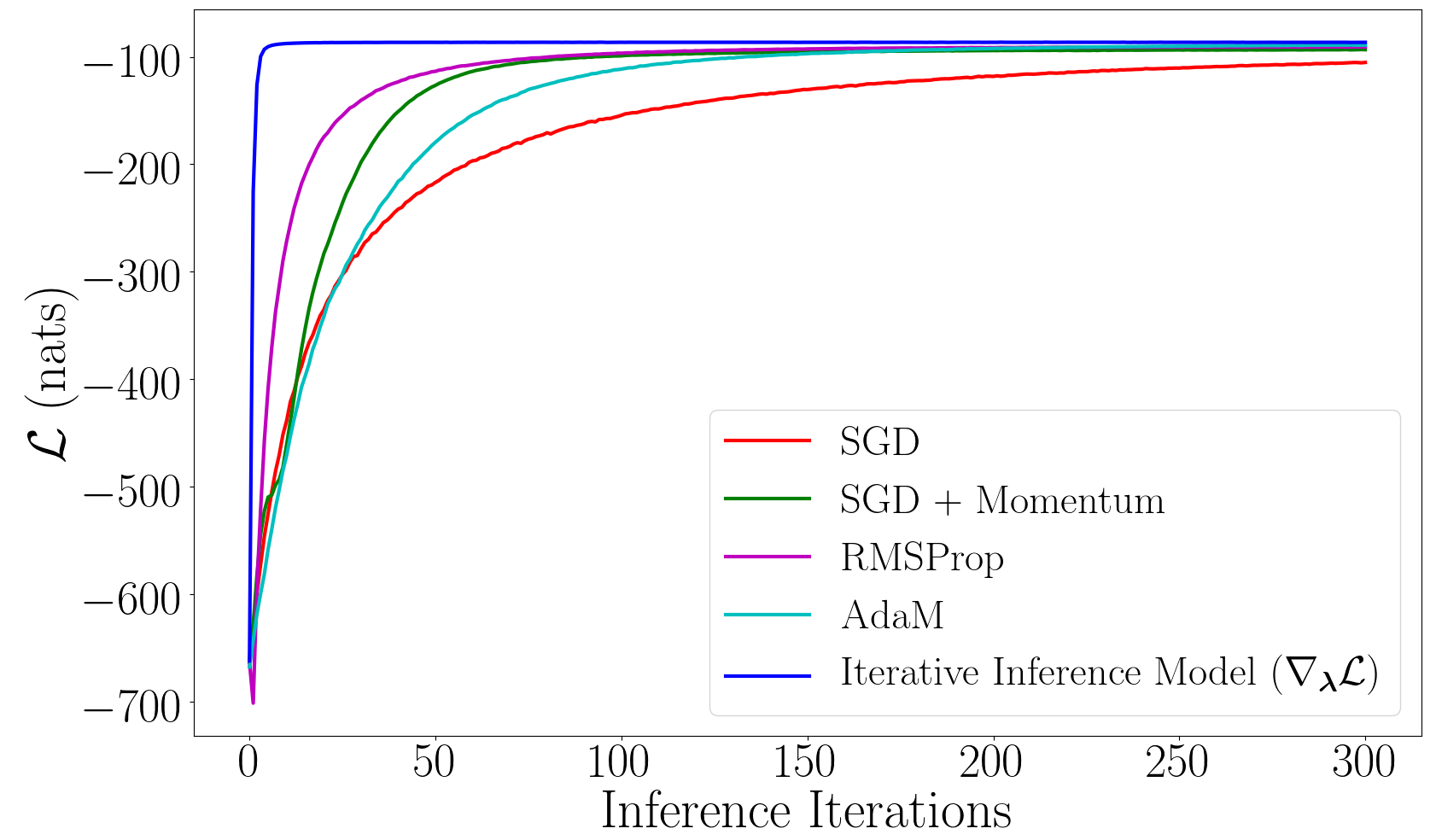}
    \captionof{figure}{\textbf{Comparison of inference optimization performance between iterative inference models and conventional optimization techniques.} Plot shows ELBO, averaged over MNIST validation set. On average, the iterative inference model converges faster than conventional optimizers to better estimates. Note that the iterative inference model remains stable over hundreds of iterations, despite only being trained with 16 inference iterations.}
    \label{fig: compare with conventional optimizers}
\end{figure}

\textbf{$\mathcal{L}$ During Inference} We can quantify and compare optimization performance through the ELBO. In Figure \ref{fig: compare with conventional optimizers}, we plot the average ELBO on the MNIST validation set during inference, comparing iterative inference models with conventional optimizers. Details are in Appendix C.2. On average, the iterative inference model converges significantly \textit{faster} to \textit{better} estimates than the optimizers. The model actually has \textit{less} derivative information than the optimizers; it only has access to the local gradient, whereas the optimizers use momentum and similar terms. The model's final estimates are also stable, despite only being trained using 16 inference iterations.

\newpage
\textbf{Reconstructions}
Approximate inference optimization can also be visualized through image reconstructions. As the reconstruction term is typically the dominant term in $\mathcal{L}$, the output reconstructions should improve in terms of visual quality during inference optimization, resembling $\mathbf{x}$. We demonstrate this phenomenon with iterative inference models for several data sets in Figure \ref{fig: reconstructions over iterations}. Additional reconstructions are shown in Appendix C.3.

\textbf{Gradient Magnitudes}
During inference optimization, iterative inference models should ideally obtain approximate posterior estimates near local maxima. The approximate posterior gradient magnitudes should thus decrease during inference. Using a model trained on RCV1, we recorded average gradient magnitudes for the approximate posterior mean during inference. In Figure \ref{fig: gradient magnitudes}, we plot these values throughout training, finding that they do, indeed, decrease. See Appendix C.4 for more details.

\begin{figure}[t!]
    \centering
    \begin{subfigure}[t]{0.36\textwidth}
        \centering
        \includegraphics[width=\textwidth]{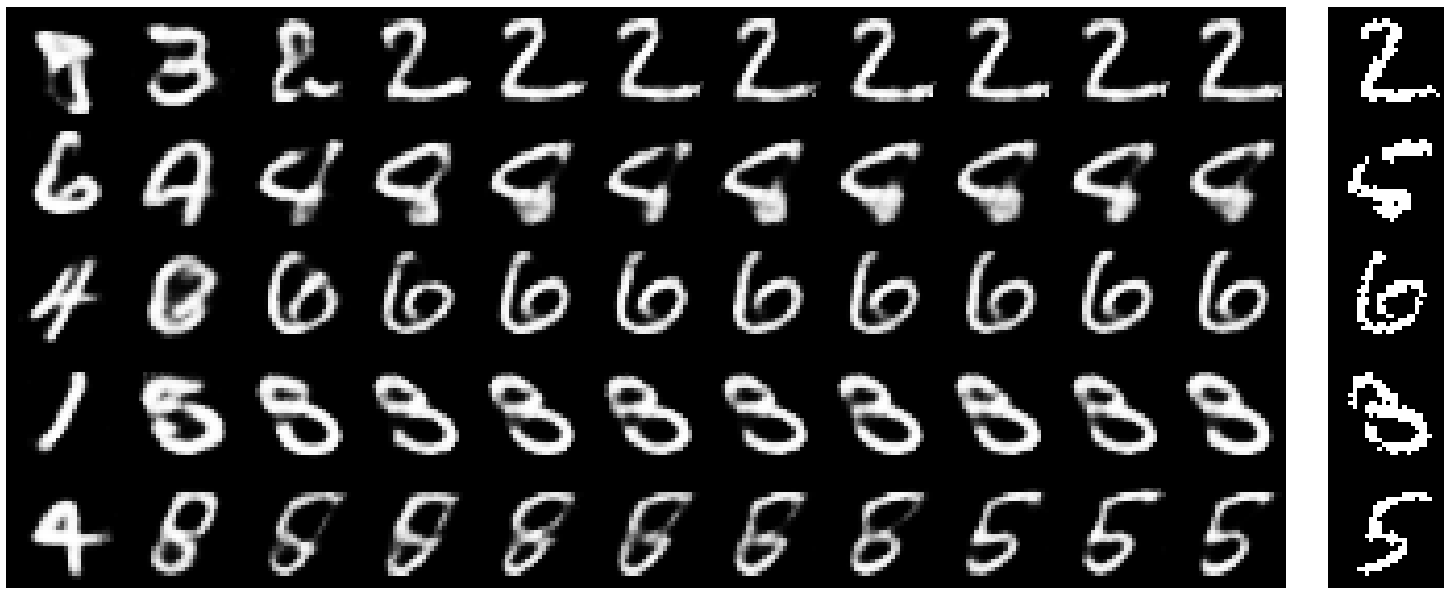}
        % \caption{}
    \end{subfigure}%
    
    \begin{subfigure}[t]{0.36\textwidth}
        \centering
        \includegraphics[width=\textwidth]{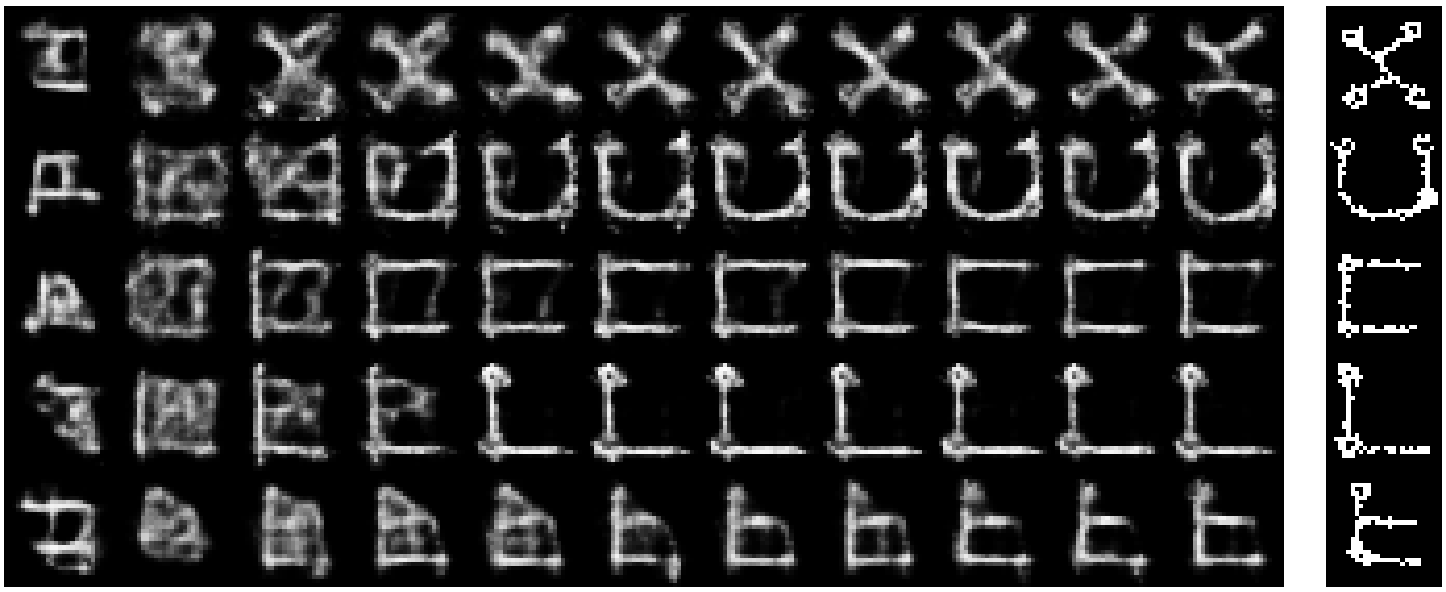}
        % \caption{}
    \end{subfigure}%
    
    \begin{subfigure}[t]{0.36\textwidth}
        \centering
        \includegraphics[width=\textwidth]{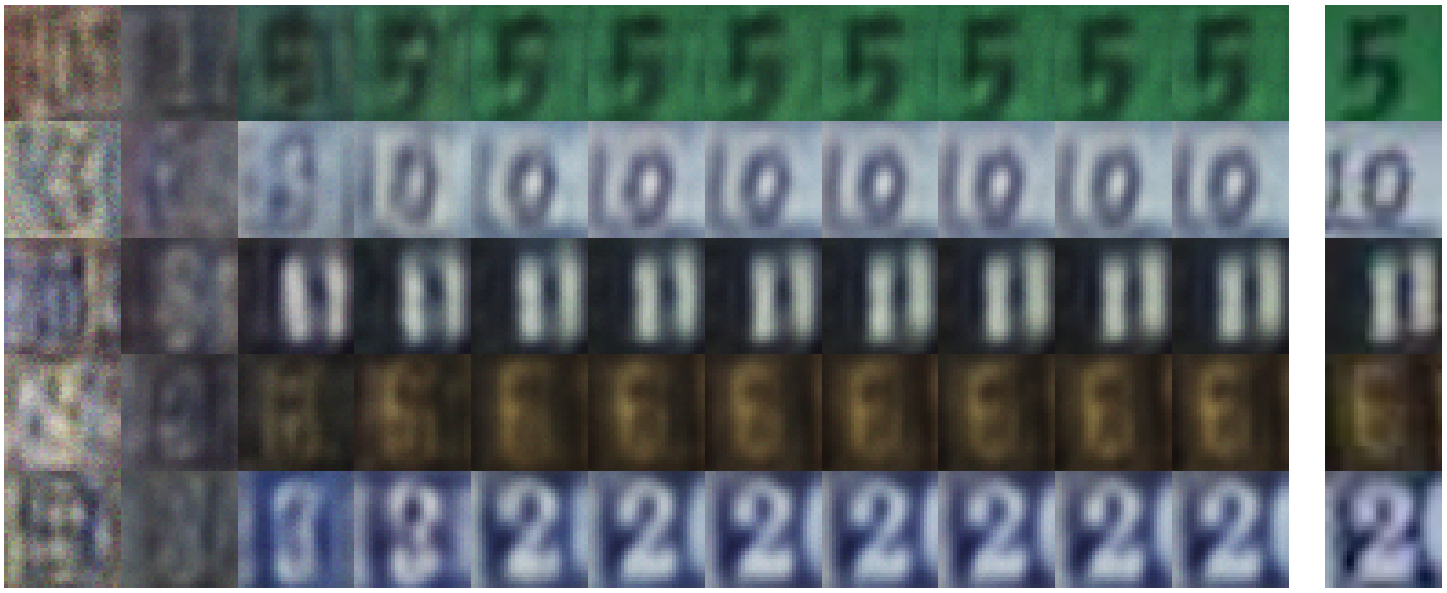}
        % \caption{}
    \end{subfigure}%
    
    \begin{subfigure}[t]{0.36\textwidth}
        \centering
        \includegraphics[width=\textwidth]{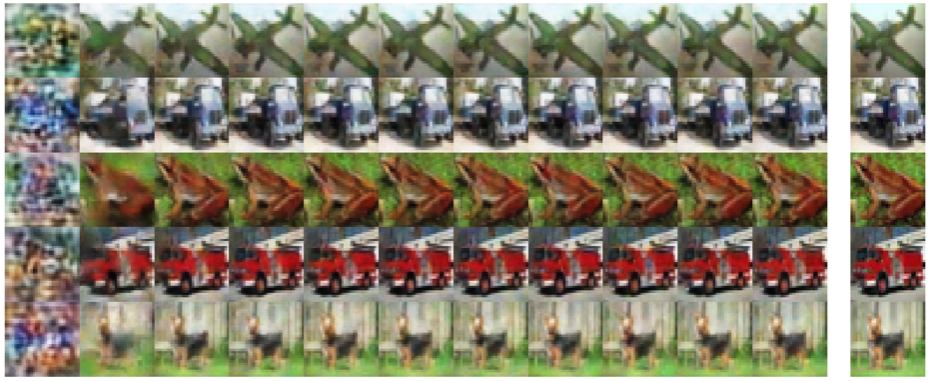}
        % \caption{}
    \end{subfigure}
    \caption{\textbf{Reconstructions over inference iterations} (left to right) for (top to bottom) MNIST, Omniglot, SVHN, and CIFAR-10. Data examples are shown on the right. Reconstructions become gradually sharper, remaining stable after many iterations.}
    \label{fig: reconstructions over iterations}
\end{figure}

\begin{figure}[t!]
    \centering        \includegraphics[width=0.48\textwidth]{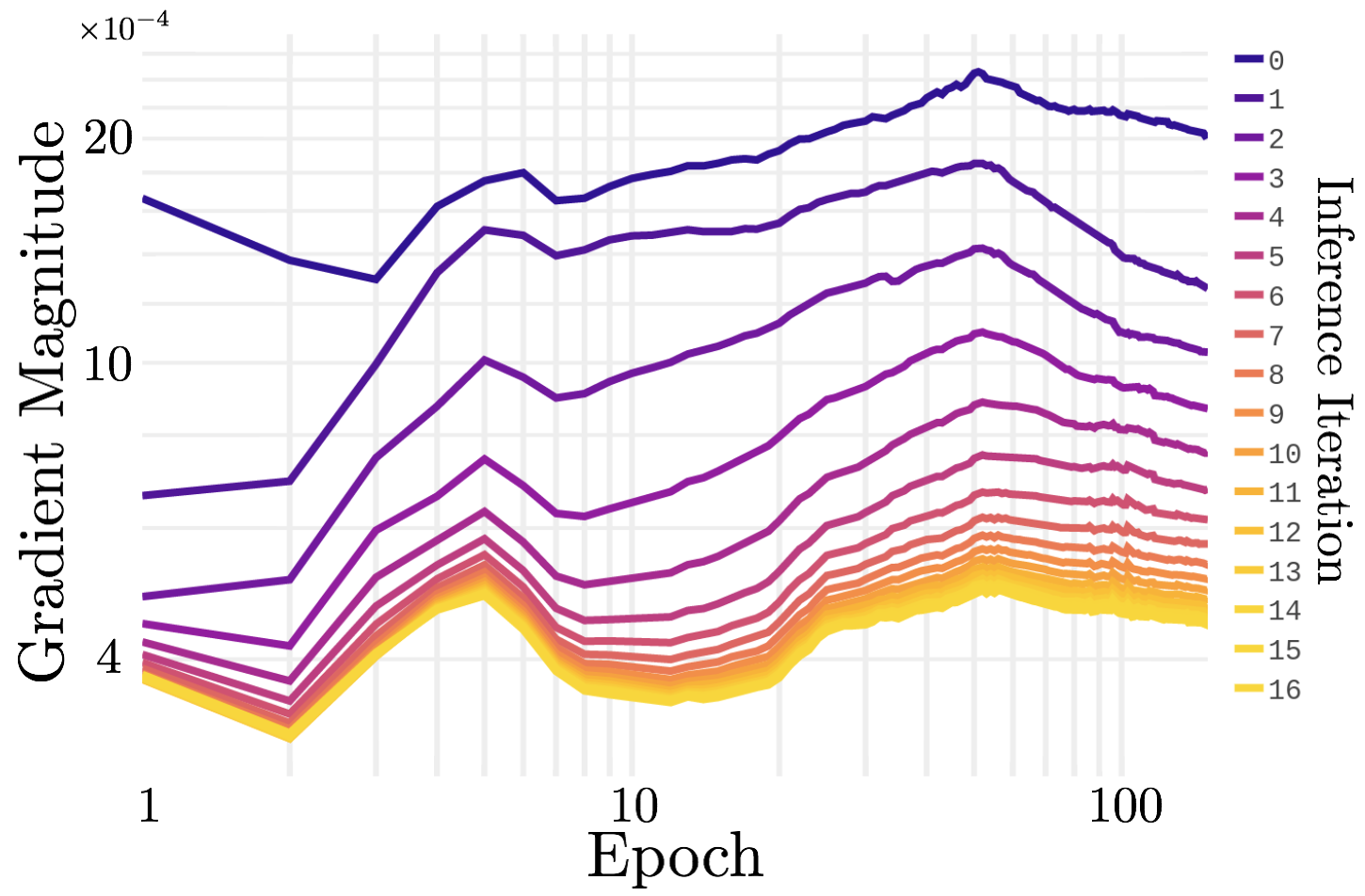}
    \caption{\textbf{Gradient magnitudes} (\textit{vertical axis}) over inference iterations (indexed by color on right) during training (\textit{horizontal axis}) on RCV1. Approx. posterior mean gradient magnitudes decrease over inference iterations as estimates approach local maxima.}
    \label{fig: gradient magnitudes}
\end{figure}

\subsection{Additional Inference Iterations \& Latent Samples}
\label{sec: samples and iterations}

\begin{figure}[t]
    \centering
    \begin{subfigure}[t]{0.45\textwidth}
        \centering
        \includegraphics[width=0.95\textwidth]{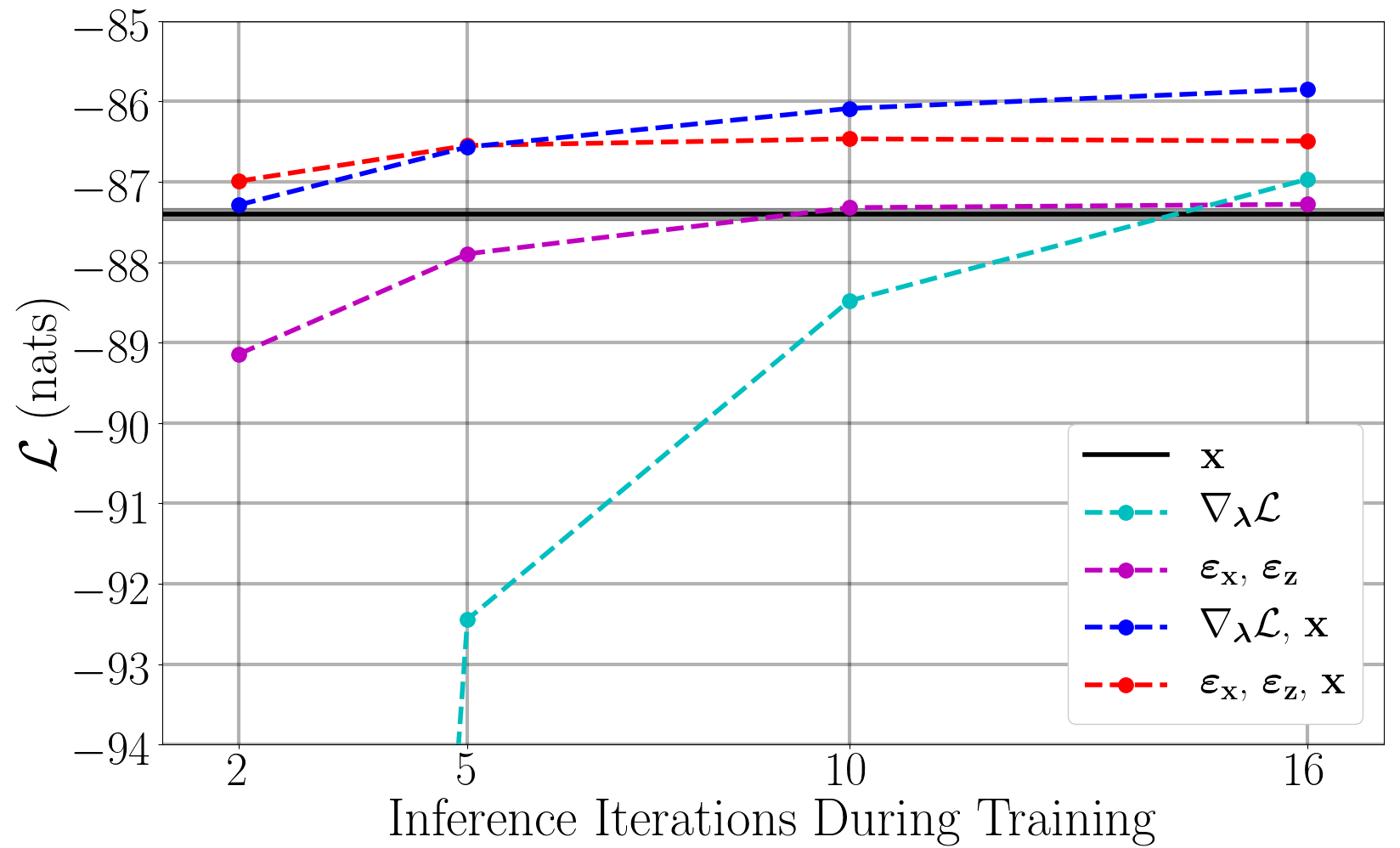}
        \caption{}
        \label{fig: additional iterations}
    \end{subfigure}
    
    \begin{subfigure}[t]{0.45\textwidth}
        \centering
        \includegraphics[width=0.97\textwidth]{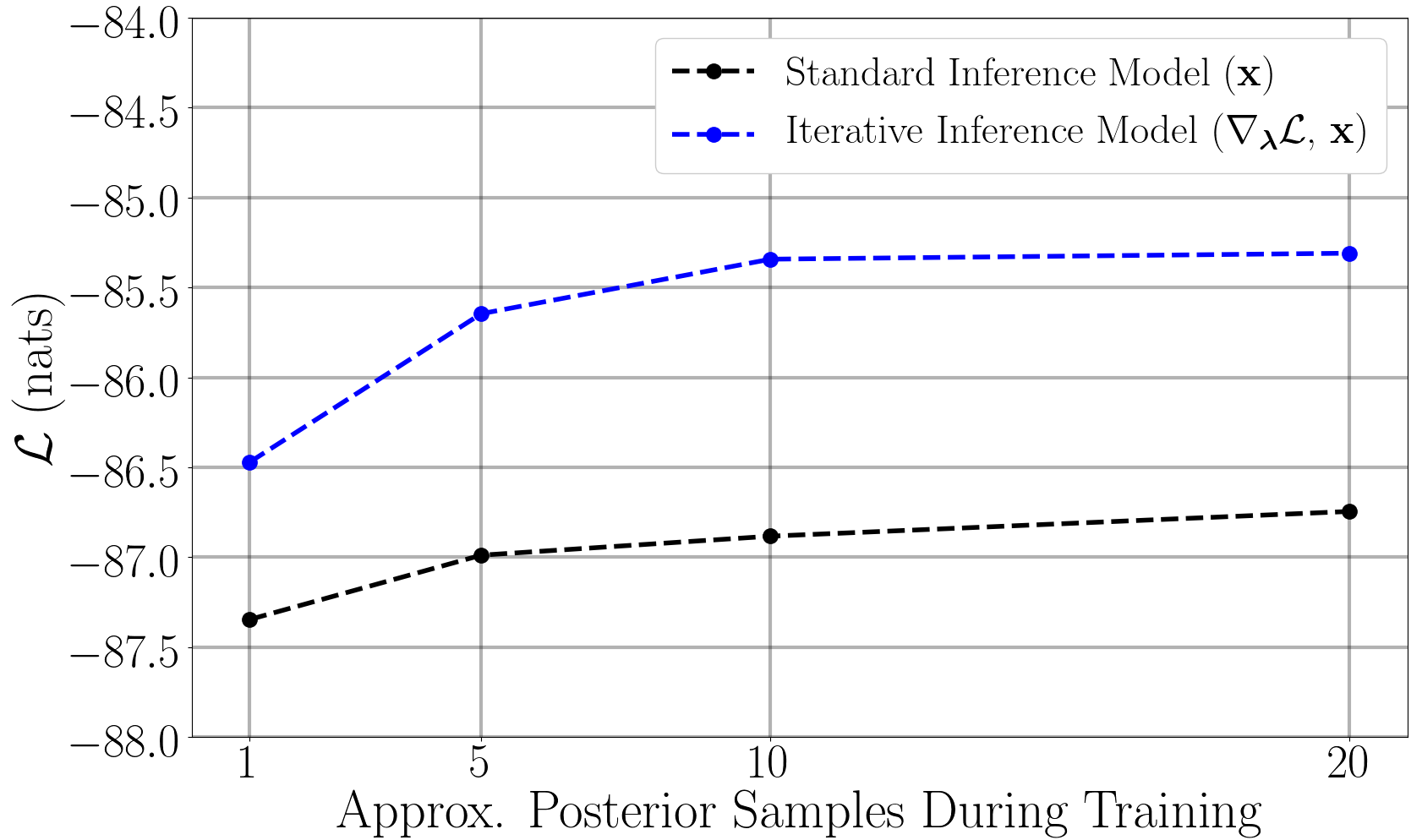}
        \caption{}
        \label{fig: additional samples}
    \end{subfigure}
    \caption{ELBO for standard and iterative inference models on MNIST for \textbf{(a)} additional inference iterations during training and \textbf{(b)} additional samples. Iterative inference models improve significantly with both quantities. Lines do not imply interpolation.}
    \label{fig: samples and iterations}
\end{figure}

We highlight two sources that allow iterative inference models to further improve modeling performance: additional inference iterations and samples. Additional inference iterations allow the model to further refine approximate posterior estimates. Using MNIST, we trained models by encoding approximate posterior gradients $(\nabla_{\bm{\lambda}} \mathcal{L})$ or errors $(\bm{\varepsilon}_\mathbf{x}, \bm{\varepsilon}_\mathbf{z})$, with or without the data $(\mathbf{x})$, for 2, 5, 10, and 16 inference iterations. While we kept the model architectures identical, the encoded terms affect the number of input parameters to each model. For instance, the small size of $\mathbf{z}$ relative to $\mathbf{x}$ gives the gradient encoding model \textit{fewer} input parameters than a standard inference model. The other models have more input parameters. Results are shown in Figure \ref{fig: additional iterations}, where we observe improved performance with increasing inference iterations. All iterative inference models outperformed standard inference models. Note that encoding errors to approximate higher-order derivatives helps when training with fewer inference iterations.

Additional approximate posterior samples provide more precise gradient and error estimates, potentially allowing an iterative inference model to output improved updates. To verify this, we trained standard and iterative inference models on MNIST using 1, 5, 10, and 20 approximate posterior samples. Iterative inference models were trained by encoding the data ($\mathbf{x}$) and approximate posterior gradients ($\nabla_{\bm{\lambda}} \mathcal{L}$) for 5 iterations. Results are shown in Figure \ref{fig: additional samples}. Iterative inference models improve by more than 1 nat with additional samples, further widening the improvement over similar standard inference models.

\subsection{Comparison with Standard Inference Models}
\label{sec: comparison with standard inference models}

We now provide a quantitative performance comparison between standard and iterative inference models on MNIST, CIFAR-10, and RCV1. Inference model architectures are identical across each comparison, with the exception of input parameters. Details are found in Appendix C.7. Table \ref{tab: quant results} contains estimated marginal log-likelihood performance on MNIST and CIFAR-10. Table \ref{tab: text results} contains estimated perplexity on RCV1\footnote{Perplexity re-weights log-likelihood by document length.}. In each case, iterative inference models outperform standard inference models. This holds for both single-level and hierarchical models. We observe larger improvements on the high-dimensional RCV1 data set, consistent with \cite{krishnan2017challenges}. Because the generative model architectures are kept fixed, performance improvements demonstrate improvements in inference optimization.

\begin{table}[t!]
\centering
\caption{Negative log likelihood on MNIST (in \textit{nats}) and CIFAR-10 (in \textit{bits/input dim.}) for standard and iterative inference models.}
\label{tab: quant results}
\centering
\begin{tabular}{l c}
\multicolumn{1}{c}{} & \multicolumn{1}{c}{$-\log p(\mathbf{x})$} \\
\hline \hline
\textbf{MNIST} \\
\hspace{2mm} \textit{Single-Level} \\
\hspace{4mm} Standard & $84.14 \pm 0.02$  \\
\hspace{4mm} Iterative & $\mathbf{83.84 \pm 0.05}$ \\
\hline
\hspace{2mm} \textit{Hierarchical} \\
\hspace{4mm} Standard & $82.63 \pm 0.01$ \\
\hspace{4mm} Iterative & $\mathbf{82.457 \pm 0.001}$ \\
\hline \hline
\textbf{CIFAR-10} \\
\hspace{2mm} \textit{Single-Level} \\
\hspace{4mm} Standard & $5.823 \pm 0.001$ \\
\hspace{4mm} Iterative &  $\mathbf{5.64 \pm 0.03}$ \\
\hline
\hspace{2mm} \textit{Hierarchical} \\
\hspace{4mm} Standard & $5.565 \pm 0.002$ \\
\hspace{4mm} Iterative & $\mathbf{5.456 \pm 0.005}$ \\
\hline
\end{tabular}
\end{table}

\begin{table}[t!]
\centering
\caption{Perplexity on RCV1 for standard and iterative inference models.}
\label{tab: text results}
\centering
\begin{tabular}{l c c}
\multicolumn{1}{c}{} & \multicolumn{1}{c}{Perplexity} & $\leq$ \\
\hline \hline
\textbf{RCV1} \\
\hspace{2mm} \citet{krishnan2017challenges} & & 331 \\
\hline
\hspace{2mm} Standard & $323 \pm 3$ & $377.4 \pm 0.5$ \\
\hspace{2mm} Iterative & $\mathbf{285.0 \pm 0.1}$ & $\mathbf{314 \pm 1}$ \\
\hline
\end{tabular}
\end{table}

\section{Conclusion}
\label{sec: conclusion}
We have proposed iterative inference models, which learn to refine inference estimates by encoding approximate posterior gradients or errors. These models generalize and extend standard inference models, and by naturally accounting for priors during inference, these models provide insight and justification for top-down inference. Through empirical evaluations, we have demonstrated that iterative inference models learn to perform variational inference optimization, with advantages over current inference techniques shown on several benchmark data sets. However, this comes with the limitation of requiring additional computation over similar standard inference models. While we discussed the relevance of iterative inference models to hierarchical latent variable models, sequential latent variable models also contain empirical priors. In future work, we hope to apply iterative inference models to the online filtering setting, where fewer inference iterations, and thus less additional computation, may be required at each time step.

\section*{Acknowledgements}

We would like to thank the reviewers as well as Peter Carr, Oisin Mac Aodha, Grant Van Horn, and Matteo Ruggero Ronchi for their insightful feedback. This research was supported in part by JPL PDF 1584398 and NSF 1564330.

\bibliography{main}
\bibliographystyle{icml2018}

\newpage
\onecolumn
\appendix

\section{Approximate Posterior Gradients for Latent Gaussian Models}
\label{appendix: gradient derivation}

\subsection{Model \& Variational Objective}

Consider a latent variable model, $p_\theta (\mathbf{x}, \mathbf{z}) = p_\theta (\mathbf{x} | \mathbf{z}) p_\theta (\mathbf{z})$, where the prior on $\mathbf{z}$ is a factorized Gaussian density, $p_\theta (\mathbf{z}) = \mathcal{N} (\mathbf{z}; \bm{\mu}_p, \diag{\bm{\sigma}^2_\mathbf{x}} )$, and the conditional likelihood, $p_\theta (\mathbf{x} | \mathbf{z})$, depends on the type of data (e.g. Bernoulli for binary observations or Gaussian for continuous observations). We introduce an approximate posterior distribution, $q (\mathbf{z} | \mathbf{x})$, which can be any parametric probability density defined over real values. Here, we assume that $q$ also takes the form of a factorized Gaussian density, $q(\mathbf{z} | \mathbf{x}) = \mathcal{N} (\mathbf{z}; \bm{\mu}_q, \diag{\bm{\sigma}^2_q} )$. The objective during variational inference is to maximize $\mathcal{L}$ w.r.t. the parameters of $q(\mathbf{z} | \mathbf{x})$, i.e. $\bm{\mu}_q$ and $\bm{\sigma}^2_q$: 
\begin{equation}
    \bm{\mu}_q^*, \bm{\sigma}_q^{2*} = \argmax_{\bm{\mu}_q, \bm{\sigma}^2_q} \mathcal{L}.
\end{equation}
To solve this optimization problem, we will use the gradients $\nabla_{\bm{\mu}_q} \mathcal{L}$ and $\nabla_{\bm{\sigma}^2_q} \mathcal{L}$, which we now derive. The objective can be written as: 
\begin{align}
    \mathcal{L} & = \mathbb{E}_{q(\mathbf{z} | \mathbf{x})} \left[ \log p_\theta (\mathbf{x}, \mathbf{z}) - \log q(\mathbf{z} | \mathbf{x}) \right] \\
    & = \mathbb{E}_{q(\mathbf{z} | \mathbf{x})} \left[ \log p_\theta (\mathbf{x} | \mathbf{z}) + \log p_\theta (\mathbf{z}) - \log q(\mathbf{z} | \mathbf{x}) \right].
\end{align}
Plugging in $p_\theta (\mathbf{z})$ and $q(\mathbf{z} | \mathbf{x})$:
\begin{equation}
    \mathcal{L} = \mathbb{E}_{\mathcal{N} (\mathbf{z}; \bm{\mu}_q, \diag{\bm{\sigma}^2_q} )} \left[ \log p_\theta (\mathbf{x} | \mathbf{z}) + \log \mathcal{N} (\mathbf{z}; \bm{\mu}_p, \diag{\bm{\sigma}^2_p} ) - \log \mathcal{N} (\mathbf{z}; \bm{\mu}_q, \diag{\bm{\sigma}^2_q} ) \right]
\end{equation}
Since expectation and differentiation are linear operators, we can take the expectation and derivative of each term individually.

\subsection{Gradient of the Log-Prior}
We can write the log-prior as:
\begin{equation}
     \log \mathcal{N} (\mathbf{z}; \bm{\mu}_p, (\diag{\bm{\sigma}^2_p} ) = -\frac{1}{2} \log \left( (2 \pi)^{n_\mathbf{z}} | \diag{\bm{\sigma}^2_p}  | \right) - \frac{1}{2} (\mathbf{z} - \bm{\mu}_p)^\intercal (\diag{\bm{\sigma}^2_p} )^{-1} (\mathbf{z} - \bm{\mu}_p),
\end{equation}
where $n_\mathbf{z}$ is the dimensionality of $\mathbf{z}$. We want to evaluate the following terms:
\begin{equation}
    \nabla_{\bm{\mu}_q} \mathbb{E}_{\mathcal{N} (\mathbf{z}; \bm{\mu}_q, \diag{\bm{\sigma}^2_q} )} \left[-\frac{1}{2} \log \left( (2 \pi)^{n_\mathbf{z}} | \diag{\bm{\sigma}^2_p}  | \right) - \frac{1}{2} (\mathbf{z} - \bm{\mu}_p)^\intercal (\diag{\bm{\sigma}^2_p} )^{-1} (\mathbf{z} - \bm{\mu}_p) \right]
    \label{eq: prior mean derivative}
\end{equation}
and
\begin{equation}
    \nabla_{\bm{\sigma}^2_q} \mathbb{E}_{\mathcal{N} (\mathbf{z}; \bm{\mu}_q, \diag{\bm{\sigma}^2_q} )} \left[-\frac{1}{2} \log \left( (2 \pi)^{n_\mathbf{z}} | \diag{\bm{\sigma}^2_p}  | \right) - \frac{1}{2} (\mathbf{z} - \bm{\mu}_p)^\intercal (\diag{\bm{\sigma}^2_p} )^{-1} (\mathbf{z} - \bm{\mu}_p) \right].
    \label{eq: prior variance derivative}
\end{equation}
To take these derivatives, we will use the reparameterization trick \cite{kingma2014stochastic, rezende2014stochastic} to re-express $\mathbf{z} = \bm{\mu}_q + \bm{\sigma}_q \odot \bm{\epsilon}$, where $\bm{\epsilon} \sim \mathcal{N}(\mathbf{0}, \mathbf{I})$ is an auxiliary standard Gaussian variable, and $\odot$ denotes the element-wise product. We can now perform the expectations over $\bm{\epsilon}$, allowing us to bring the gradient operators inside the expectation brackets. The first term in eqs. \ref{eq: prior mean derivative} and \ref{eq: prior variance derivative} does not depend on $\bm{\mu}_q$ or $\bm{\sigma}^2_q$, so we can write:
\begin{equation}
    \mathbb{E}_{\mathcal{N} (\bm{\epsilon}; \mathbf{0}, \mathbf{I})} \left[\nabla_{\bm{\mu}_q} \left( - \frac{1}{2} (\bm{\mu}_q + \bm{\sigma}_q \odot \bm{\epsilon} - \bm{\mu}_p)^\intercal (\diag{\bm{\sigma}^2_p} )^{-1} (\bm{\mu}_q + \bm{\sigma}_q \odot \bm{\epsilon} - \bm{\mu}_p) \right) \right]
    \label{eq: prior mean derivative 2}
\end{equation}
and
\begin{equation}
     \mathbb{E}_{\mathcal{N} (\bm{\epsilon}; \mathbf{0}, \mathbf{I})} \left[ \nabla_{\bm{\sigma}^2_q} \left( - \frac{1}{2} (\bm{\mu}_q + \bm{\sigma}_q \odot \bm{\epsilon} - \bm{\mu}_p)^\intercal (\diag{\bm{\sigma}^2_p} )^{-1} (\bm{\mu}_q + \bm{\sigma}_q \odot \bm{\epsilon} - \bm{\mu}_p) \right) \right].
    \label{eq: prior variance derivative 2}
\end{equation}
To simplify notation, we define the following term:
\begin{equation}
\bm{\xi} \equiv (\diag{\bm{\sigma}^2_p} )^{-1/2} (\bm{\mu}_q + \bm{\sigma}_q \odot \bm{\epsilon} - \bm{\mu}_p),
\end{equation}
allowing us to rewrite eqs. \ref{eq: prior mean derivative 2} and \ref{eq: prior variance derivative 2} as:
\begin{equation}
    \mathbb{E}_{\mathcal{N} (\bm{\epsilon}; \mathbf{0}, \mathbf{I})} \left[\nabla_{\bm{\mu}_q} \left( - \frac{1}{2} \bm{\xi}^\intercal \bm{\xi} \right) \right] = \mathbb{E}_{\mathcal{N} (\bm{\epsilon}; \mathbf{0}, \mathbf{I})} \left[ - \frac{\partial \bm{\xi}^\intercal}{\partial \bm{\mu}_q} \bm{\xi} \right]
    \label{eq: prior mean derivative 3}
\end{equation}
and
\begin{equation}
     \mathbb{E}_{\mathcal{N} (\bm{\epsilon}; \mathbf{0}, \mathbf{I})} \left[ \nabla_{\bm{\sigma}^2_q} \left( - \frac{1}{2} \bm{\xi}^\intercal \bm{\xi} \right) \right] = \mathbb{E}_{\mathcal{N} (\bm{\epsilon}; \mathbf{0}, \mathbf{I})} \left[ - \frac{\partial \bm{\xi}^\intercal}{\partial \bm{\sigma}^2_q} \bm{\xi} \right].
    \label{eq: prior variance derivative 3}
\end{equation}
We must now find $\frac{\partial \bm{\xi}}{\partial \bm{\mu}_q}$ and $\frac{\partial \bm{\xi}}{\partial \bm{\sigma}^2_q}$:
\begin{equation}
    \frac{\partial \bm{\xi}}{\partial \bm{\mu}_q} = \frac{\partial}{\partial \bm{\mu}_q} \left( (\diag{\bm{\sigma}^2_p} )^{-1/2} (\bm{\mu}_q + \bm{\sigma}_q \odot \bm{\epsilon} - \bm{\mu}_p) \right) = (\diag{\bm{\sigma}^2_p} )^{-1/2}
    \label{eq: prior mean derivative 4}
\end{equation}
and
\begin{equation}
    \frac{\partial \bm{\xi}}{\partial \bm{\sigma}^2_q} = \frac{\partial}{\partial \bm{\sigma}^2_q} \left( (\diag{\bm{\sigma}^2_p} )^{-1/2} (\bm{\mu}_q + \bm{\sigma}_q \odot \bm{\epsilon} - \bm{\mu}_p) \right) = (\diag{\bm{\sigma}^2_p} )^{-1/2} \diag{\frac{\bm{\epsilon}}{2 \bm{\sigma}_q}},
    \label{eq: prior variance derivative 4}
\end{equation}
where division is performed element-wise. Plugging eqs. \ref{eq: prior mean derivative 4} and \ref{eq: prior variance derivative 4} back into eqs. \ref{eq: prior mean derivative 3} and \ref{eq: prior variance derivative 3}, we get:
\begin{equation}
    \mathbb{E}_{\mathcal{N} (\bm{\epsilon}; \mathbf{0}, \mathbf{I})} \left[ - \left((\diag{\bm{\sigma}^2_p} )^{-1/2}\right)^\intercal (\diag{\bm{\sigma}^2_p} )^{-1/2} (\bm{\mu}_q + \bm{\sigma}_q \odot \bm{\epsilon} - \bm{\mu}_p) \right] 
\end{equation}
and
\begin{equation}
    \mathbb{E}_{\mathcal{N} (\bm{\epsilon}; \mathbf{0}, \mathbf{I})} \left[ - \left( \diag{\frac{\bm{\epsilon}}{2 \bm{\sigma}_q}} \right)^\intercal \left((\diag{\bm{\sigma}^2_p} )^{-1/2}\right)^\intercal (\diag{\bm{\sigma}^2_p} )^{-1/2} (\bm{\mu}_q + \bm{\sigma}_q \odot \bm{\epsilon} - \bm{\mu}_p) \right] .
\end{equation}
Putting everything together, we can express the gradients as:
\begin{equation}
    \nabla_{\bm{\mu}_q} \mathbb{E}_{\mathcal{N} (\mathbf{z}; \bm{\mu}_q, \diag{\bm{\sigma}^2_q} )} \left[ \log \mathcal{N} (\mathbf{z}; \bm{\mu}_p, \diag{\bm{\sigma}^2_p} ) \right] = \mathbb{E}_{\mathcal{N} (\bm{\epsilon}; \mathbf{0}, \mathbf{I} )} \left[ -\frac{\bm{\mu}_q + \bm{\sigma}_q \odot \bm{\epsilon} - \bm{\mu}_p}{\bm{\sigma}^2_p} \right],
\end{equation}
and
\begin{equation}
    \nabla_{\bm{\sigma}^2_q} \mathbb{E}_{\mathcal{N} (\mathbf{z}; \bm{\mu}_q, \diag{\bm{\sigma}^2_q} )} \left[ \log \mathcal{N} (\mathbf{z}; \bm{\mu}_p, \diag{\bm{\sigma}^2_p} ) \right] = \mathbb{E}_{\mathcal{N} (\bm{\epsilon}; \mathbf{0}, \mathbf{I} )} \left[- \left( \diag{\frac{\bm{\epsilon}}{2 \bm{\sigma}_q}} \right)^\intercal \frac{\bm{\mu}_q + \bm{\sigma}_q \odot \bm{\epsilon} - \bm{\mu}_p}{\bm{\sigma}^2_p} \right].
\end{equation}

\subsection{Gradient of the Log-Approximate Posterior}
We can write the log-approximate posterior as:
\begin{equation}
     \log \mathcal{N} (\mathbf{z}; \bm{\mu}_q, \diag{\bm{\sigma}^2_q} ) = -\frac{1}{2} \log \left( (2 \pi)^{n_\mathbf{z}} | \diag{\bm{\sigma}^2_q}  | \right) - \frac{1}{2} (\mathbf{z} - \bm{\mu}_q)^\intercal (\diag{\bm{\sigma}^2_q} )^{-1} (\mathbf{z} - \bm{\mu}_q),
     \label{eq: log approximate posterior}
\end{equation}
where $n_\mathbf{z}$ is the dimensionality of $\mathbf{z}$. Again, we will use the reparameterization trick to re-express the gradients. However, notice what happens when plugging the reparameterized $\mathbf{z} = \bm{\mu}_q + \bm{\sigma}_q \odot \bm{\epsilon}$ into the second term of eq. \ref{eq: log approximate posterior}:
\begin{equation}
    - \frac{1}{2} (\bm{\mu}_q + \bm{\sigma}_q \odot \bm{\epsilon} - \bm{\mu}_q)^\intercal (\diag{\bm{\sigma}^2_q} )^{-1} (\bm{\mu}_q + \bm{\sigma}_q \odot \bm{\epsilon} - \bm{\mu}_q) = - \frac{1}{2} \frac{(\bm{\sigma}_q \odot \bm{\epsilon})^\intercal (\bm{\sigma}_q \odot \bm{\epsilon}) }{\bm{\sigma}^2_q} = - \frac{1}{2}\bm{\epsilon}^\intercal \bm{\epsilon}.
\end{equation}
This term does not depend on $\bm{\mu}_q$ or $\bm{\sigma}^2_q$. Also notice that the first term in eq. \ref{eq: log approximate posterior} depends only on $\bm{\sigma}^2_q$. Therefore, the gradient of the entire term w.r.t. $\bm{\mu}_q$ is zero:
\begin{equation}
    \nabla_{\bm{\mu}_q} \mathbb{E}_{\mathcal{N} (\mathbf{z}; \bm{\mu}_q, \diag{\bm{\sigma}^2_q} )} \left[ \log \mathcal{N} (\mathbf{z}; \bm{\mu}_q, \diag{\bm{\sigma}^2_q} ) \right] = \mathbf{0}.
    \label{eq: posterior mean gradient 1}
\end{equation}
The gradient w.r.t. $\bm{\sigma}^2_q$ is
\begin{equation}
     \nabla_{\bm{\sigma}^2_q} \left(-\frac{1}{2} \log \left( (2 \pi)^{n_\mathbf{z}} | \diag{\bm{\sigma}^2_q}  | \right) \right) = -\frac{1}{2} \nabla_{\bm{\sigma}^2_q} \left( \log | \diag{\bm{\sigma}^2_q}  | \right) = -\frac{1}{2} \nabla_{\bm{\sigma}^2_q} \sum_j \log \sigma_{q, j}^2 = -\frac{\mathbf{1}}{2 \bm{\sigma}^2_q}.
    \label{eq: posterior variance derivative 1}
\end{equation}
Note that the expectation has been dropped, as the term does not depend on the value of the sampled $\mathbf{z}$. Thus, the gradient of the entire term w.r.t. $\bm{\sigma}^2_q$ is:
\begin{equation}
    \nabla_{\bm{\sigma}^2_q} \mathbb{E}_{\mathcal{N} (\mathbf{z}; \bm{\mu}_q, \diag{\bm{\sigma}^2_q} )} \left[ \log \mathcal{N} (\mathbf{z}; \bm{\mu}_q, \diag{\bm{\sigma}^2_q} ) \right] = -\frac{\mathbf{1}}{2 \bm{\sigma}^2_q}.
    \label{eq: posterior variance gradient 2}
\end{equation}

\subsection{Gradient of the Log-Conditional Likelihood}
The form of the conditional likelihood will depend on the data, e.g. binary, discrete, continuous, etc. Here, we derive the gradient for Bernoulli (binary) and Gaussian (continuous) conditional likelihoods.

\paragraph{Bernoulli Output Distribution}
The $\log$ of a Bernoulli output distribution takes the form:
\begin{equation}
    \log \mathcal{B} (\mathbf{x} ; \bm{\mu}_\mathbf{x}) = (\log \bm{\mu}_\mathbf{x})^\intercal  \mathbf{x} +  (\log (\mathbf{1} - \bm{\mu}_\mathbf{x})) ^\intercal (\mathbf{1} - \mathbf{x}),
\end{equation}
where $\bm{\mu}_\mathbf{x} = \bm{\mu}_\mathbf{x} (\mathbf{z}, \theta)$ is the mean of the output distribution. We drop the explicit dependence on $\mathbf{z}$ and $\theta$ to simplify notation. We want to compute the gradients
\begin{equation}
    \nabla_{\bm{\mu}_q} \mathbb{E}_{\mathcal{N}(\mathbf{z}; \bm{\mu}_q, \diag{\bm{\sigma}^2_q} )} \left[ (\log \bm{\mu}_\mathbf{x})^\intercal  \mathbf{x} +  (\log (\mathbf{1} - \bm{\mu}_\mathbf{x})) ^\intercal (\mathbf{1} - \mathbf{x}) \right]
    \label{eq: cond like mean gradient 1}
\end{equation}
and
\begin{equation}
    \nabla_{\bm{\sigma}^2_q} \mathbb{E}_{\mathcal{N}(\mathbf{z}; \bm{\mu}_q, \diag{\bm{\sigma}^2_q} )} \left[ (\log \bm{\mu}_\mathbf{x})^\intercal  \mathbf{x} +  (\log (\mathbf{1} - \bm{\mu}_\mathbf{x})) ^\intercal (\mathbf{1} - \mathbf{x}) \right].
    \label{eq: cond like variance gradient 1}
\end{equation}
Again, we use the reparameterization trick to re-express the expectations, allowing us to bring the gradient operators inside the brackets. Using $\mathbf{z} = \bm{\mu}_q + \bm{\sigma}_q \odot \bm{\epsilon}$, eqs. \ref{eq: cond like mean gradient 1} and \ref{eq: cond like variance gradient 1} become:
\begin{equation}
     \mathbb{E}_{\mathcal{N}(\bm{\epsilon}; \mathbf{0}, \mathbf{I})} \left[ \nabla_{\bm{\mu}_q} \left( (\log \bm{\mu}_\mathbf{x})^\intercal  \mathbf{x} +  (\log (\mathbf{1} - \bm{\mu}_\mathbf{x})) ^\intercal (\mathbf{1} - \mathbf{x}) \right) \right]
    \label{eq: cond like mean gradient 2}
\end{equation}
and
\begin{equation}
     \mathbb{E}_{\mathcal{N}(\bm{\epsilon}; \mathbf{0}, \mathbf{I})} \left[ \nabla_{\bm{\sigma}^2_q} \left( (\log \bm{\mu}_\mathbf{x})^\intercal  \mathbf{x} +  (\log (\mathbf{1} - \bm{\mu}_\mathbf{x})) ^\intercal (\mathbf{1} - \mathbf{x}) \right) \right],
    \label{eq: cond like variance gradient 2}
\end{equation}
where $\bm{\mu}_\mathbf{x}$ is re-expressed as function of $\bm{\mu}_q, \bm{\sigma}^2_q, \bm{\epsilon},$ and $\theta$. Distributing the gradient operators yields:
\begin{equation}
     \mathbb{E}_{\mathcal{N}(\bm{\epsilon}; \mathbf{0}, \mathbf{I})} \left[ \frac{\partial (\log \bm{\mu}_\mathbf{x})^\intercal}{\partial \bm{\mu}_q} \mathbf{x} +  \frac{\partial (\log (\mathbf{1} - \bm{\mu}_\mathbf{x}))^\intercal}{\partial \bm{\mu}_q} (\mathbf{1} - \mathbf{x}) \right]
    \label{eq: cond like mean gradient 3}
\end{equation}
and
\begin{equation}
     \mathbb{E}_{\mathcal{N}(\bm{\epsilon}; \mathbf{0}, \mathbf{I})} \left[ \frac{\partial (\log \bm{\mu}_\mathbf{x})^\intercal}{\partial \bm{\sigma}^2_q} \mathbf{x} +  \frac{\partial (\log (\mathbf{1} - \bm{\mu}_\mathbf{x}))^\intercal}{\partial \bm{\sigma}^2_q} (\mathbf{1} - \mathbf{x}) \right].
    \label{eq: cond like variance gradient 3}
\end{equation}
Taking the partial derivatives and combining terms gives:
\begin{equation}
     \mathbb{E}_{\mathcal{N}(\bm{\epsilon}; \mathbf{0}, \mathbf{I})} \left[ {\frac{\partial \bm{\mu}_\mathbf{x}}{\partial \bm{\mu}_q}}^\intercal \frac{\mathbf{x}}{\bm{\mu}_\mathbf{x}} - {\frac{\partial \bm{\mu}_\mathbf{x}}{\partial \bm{\mu}_q}}^\intercal \frac{\mathbf{1} - \mathbf{x}}{\mathbf{1} - \bm{\mu}_\mathbf{x}} \right] = \mathbb{E}_{\mathcal{N}(\bm{\epsilon}; \mathbf{0}, \mathbf{I})} \left[ {\frac{\partial \bm{\mu}_\mathbf{x}}{\partial \bm{\mu}_q}}^\intercal \frac{\mathbf{x} - \bm{\mu}_\mathbf{x}}{\bm{\mu}_\mathbf{x} \odot (\mathbf{1} - \bm{\mu}_\mathbf{x})} \right]
    \label{eq: cond like mean gradient 4}
\end{equation}
and
\begin{equation}
     \mathbb{E}_{\mathcal{N}(\bm{\epsilon}; \mathbf{0}, \mathbf{I})} \left[ {\frac{\partial \bm{\mu}_\mathbf{x}}{\partial \bm{\sigma}^2_q}}^\intercal \frac{\mathbf{x}}{\bm{\mu}_\mathbf{x}} - {\frac{\partial \bm{\mu}_\mathbf{x}}{\partial \bm{\sigma}^2_q}}^\intercal \frac{\mathbf{1} - \mathbf{x}}{\mathbf{1} - \bm{\mu}_\mathbf{x}} \right] = \mathbb{E}_{\mathcal{N}(\bm{\epsilon}; \mathbf{0}, \mathbf{I})} \left[ {\frac{\partial \bm{\mu}_\mathbf{x}}{\partial \bm{\sigma}^2_q}}^\intercal \frac{\mathbf{x} - \bm{\mu}_\mathbf{x}}{\bm{\mu}_\mathbf{x} \odot (\mathbf{1} - \bm{\mu}_\mathbf{x})} \right].
    \label{eq: cond like variance gradient 4}
\end{equation}

\paragraph{Gaussian Output Density}

The $\log$ of a Gaussian output density takes the form:
\begin{equation}
     \log \mathcal{N} (\mathbf{x}; \bm{\mu}_\mathbf{x}, \diag{\bm{\sigma}^2_\mathbf{x}}) = -\frac{1}{2} \log \left( (2 \pi)^{n_\mathbf{x}} | \diag{\bm{\sigma}^2_\mathbf{x}} | \right) - \frac{1}{2} (\mathbf{x} - \bm{\mu}_\mathbf{x})^\intercal (\diag{\bm{\sigma}^2_\mathbf{x}})^{-1} (\mathbf{x} - \bm{\mu}_\mathbf{x}),
     \label{eq: log gaussian output}
\end{equation}
where $\bm{\mu}_\mathbf{x} = \bm{\mu}_\mathbf{x} (\mathbf{z}, \theta)$ is the mean of the output distribution and $\bm{\sigma}^2_\mathbf{x} = \bm{\sigma}^2_\mathbf{x} (\theta)$ is the variance. We assume $\bm{\sigma}^2_\mathbf{x}$ is not a function of $\mathbf{z}$ to simplify the derivation, however, using $\bm{\sigma}^2_\mathbf{x} = \bm{\sigma}^2_\mathbf{x} (\mathbf{z}, \theta)$ is possible and would simply result in additional gradient terms in $\nabla_{\bm{\mu}_q} \mathcal{L}$ and $\nabla_{\bm{\sigma}^2_q} \mathcal{L}$. We want to compute the gradients
\begin{equation}
    \nabla_{\bm{\mu}_q} \mathbb{E}_{\mathcal{N} (\mathbf{z}; \bm{\mu}_q, \diag{\bm{\sigma}^2_q} )} \left[-\frac{1}{2} \log \left( (2 \pi)^{n_\mathbf{x}} | \diag{\bm{\sigma}^2_\mathbf{x}} | \right) - \frac{1}{2} (\mathbf{x} - \bm{\mu}_\mathbf{x})^\intercal (\diag{\bm{\sigma}^2_\mathbf{x}})^{-1} (\mathbf{x} - \bm{\mu}_\mathbf{x}) \right]
    \label{eq: cond like mean derivative}
\end{equation}
and
\begin{equation}
    \nabla_{\bm{\sigma}^2_q} \mathbb{E}_{\mathcal{N} (\mathbf{z}; \bm{\mu}_q, \diag{\bm{\sigma}^2_q} )} \left[-\frac{1}{2} \log \left( (2 \pi)^{n_\mathbf{x}} | \diag{\bm{\sigma}^2_\mathbf{x}} | \right) - \frac{1}{2} (\mathbf{x} - \bm{\mu}_\mathbf{x})^\intercal (\diag{\bm{\sigma}^2_\mathbf{x}})^{-1} (\mathbf{x} - \bm{\mu}_\mathbf{x}) \right].
    \label{eq: cond like variance derivative}
\end{equation}
The first term in eqs. \ref{eq: cond like mean derivative} and \ref{eq: cond like variance derivative} is zero, since $\bm{\sigma}^2_\mathbf{x}$ does not depend on $\bm{\mu}_q$ or $\bm{\sigma}^2_q$. To take the gradients, we will again use the reparameterization trick to re-express $\mathbf{z} = \bm{\mu}_q + \bm{\sigma}_q \odot \bm{\epsilon}$. We now implicitly express $\bm{\mu}_\mathbf{x}$ as   $\bm{\mu}_\mathbf{x} (\bm{\mu}_q, \bm{\sigma}^2_q, \theta)$. We can then write:
\begin{equation}
     \mathbb{E}_{\mathcal{N} (\bm{\epsilon}; \mathbf{0}, \mathbf{I})} \left[\nabla_{\bm{\mu}_q} \left( - \frac{1}{2} (\mathbf{x} - \bm{\mu}_\mathbf{x})^\intercal (\diag{\bm{\sigma}^2_\mathbf{x}})^{-1} (\mathbf{x} - \bm{\mu}_\mathbf{x}) \right) \right]
    \label{eq: cond like mean derivative 2}
\end{equation}
and
\begin{equation}
     \mathbb{E}_{\mathcal{N} (\bm{\epsilon}; \mathbf{0}, \mathbf{I})} \left[ \nabla_{\bm{\sigma}^2_q} \left( - \frac{1}{2} (\mathbf{x} - \bm{\mu}_\mathbf{x})^\intercal (\diag{\bm{\sigma}^2_\mathbf{x}})^{-1} (\mathbf{x} - \bm{\mu}_\mathbf{x}) \right) \right].
    \label{eq: cond like variance derivative 2}
\end{equation}
To simplify notation, we define the following term:
\begin{equation}
\bm{\xi} \equiv (\diag{\bm{\sigma}^2_\mathbf{x}})^{-1/2} (\mathbf{x} - \bm{\mu}_\mathbf{x}),
\end{equation}
allowing us to rewrite eqs. \ref{eq: cond like mean derivative 2} and \ref{eq: cond like variance derivative 2} as 
\begin{equation}
     \mathbb{E}_{\mathcal{N} (\bm{\epsilon}; \mathbf{0}, \mathbf{I})} \left[\nabla_{\bm{\mu}_q} \left( - \frac{1}{2} \bm{\xi}^\intercal \bm{\xi} \right) \right] = \mathbb{E}_{\mathcal{N} (\bm{\epsilon}; \mathbf{0}, \mathbf{I})} \left[ - \frac{\partial \bm{\xi}^\intercal }{\partial \bm{\mu}_q} \bm{\xi} \right]
    \label{eq: cond like mean derivative 3}
\end{equation}
and
\begin{equation}
     \mathbb{E}_{\mathcal{N} (\bm{\epsilon}; \mathbf{0}, \mathbf{I})} \left[ \nabla_{\bm{\sigma}^2_q} \left( - \frac{1}{2} \bm{\xi}^\intercal \bm{\xi} \right) \right] = \mathbb{E}_{\mathcal{N} (\bm{\epsilon}; \mathbf{0}, \mathbf{I})} \left[ - \frac{\partial \bm{\xi}^\intercal }{\partial \bm{\sigma}^2_q} \bm{\xi} \right].
    \label{eq: cond like variance derivative 3}
\end{equation}
We must now find $\frac{\partial \bm{\xi}}{\partial \bm{\mu}_q}$ and $\frac{\partial \bm{\xi}}{\partial \bm{\sigma}^2_q}$:
\begin{equation}
    \frac{\partial \bm{\xi}}{\partial \bm{\mu}_q} = \frac{\partial}{\partial \bm{\mu}_q} \left( (\diag{\bm{\sigma}^2_\mathbf{x}})^{-1/2} (\mathbf{x} - \bm{\mu}_\mathbf{x}) \right) = - (\diag{\bm{\sigma}^2_\mathbf{x}})^{-1/2} \frac{\partial \bm{\mu}_\mathbf{x}}{\partial \bm{\mu}_q}
\end{equation}
and
\begin{equation}
    \frac{\partial \bm{\xi}}{\partial \bm{\sigma}^2_q} = \frac{\partial}{\partial \bm{\sigma}^2_q} \left( (\diag{\bm{\sigma}^2_\mathbf{x}})^{-1/2} (\mathbf{x} - \bm{\mu}_\mathbf{x}) \right) = - (\diag{\bm{\sigma}^2_\mathbf{x}})^{-1/2} \frac{\partial \bm{\mu}_\mathbf{x}}{\partial \bm{\sigma}^2_q}.
\end{equation}
Plugging these expressions back into eqs. \ref{eq: cond like mean derivative 3} and \ref{eq: cond like variance derivative 3} gives
\begin{equation}
     \mathbb{E}_{\mathcal{N} (\bm{\epsilon}; \mathbf{0}, \mathbf{I})} \left[ {\frac{\partial \bm{\mu}_\mathbf{x}}{\partial \bm{\mu}_q}}^\intercal ((\diag{\bm{\sigma}^2_\mathbf{x}})^{-1/2})^\intercal (\diag{\bm{\sigma}^2_\mathbf{x}})^{-1/2} (\mathbf{x} - \bm{\mu}_\mathbf{x}) \right] = \mathbb{E}_{\mathcal{N} (\bm{\epsilon}; \mathbf{0}, \mathbf{I})} \left[ {\frac{\partial \bm{\mu}_\mathbf{x} }{\partial \bm{\mu}_q}}^\intercal \frac{\mathbf{x} - \bm{\mu}_\mathbf{x}}{\bm{\sigma}_\mathbf{x}^2} \right]
    \label{eq: cond like mean derivative 4}
\end{equation}
and
\begin{equation}
      \mathbb{E}_{\mathcal{N} (\bm{\epsilon}; \mathbf{0}, \mathbf{I})} \left[ {\frac{\partial \bm{\mu}_\mathbf{x}}{\partial \bm{\sigma}^2_q}}^\intercal ((\diag{\bm{\sigma}^2_\mathbf{x}})^{-1/2})^\intercal (\diag{\bm{\sigma}^2_\mathbf{x}})^{-1/2} (\mathbf{x} - \bm{\mu}_\mathbf{x}) \right] = \mathbb{E}_{\mathcal{N} (\bm{\epsilon}; \mathbf{0}, \mathbf{I})} \left[ {\frac{\partial \bm{\mu}_\mathbf{x}}{\partial \bm{\sigma}^2_q}}^\intercal \frac{\mathbf{x} - \bm{\mu}_\mathbf{x}}{\bm{\sigma}_\mathbf{x}^2} \right].
    \label{eq: cond like variance derivative 4}
\end{equation}

Despite having different distribution forms, Bernoulli and Gaussian output distributions result in approximate posterior gradients of a similar form: the Jacobian of the output model multiplied by a weighted error term.

\subsection{Summary}
Putting the gradient terms from $\log p_\theta (\mathbf{x} | \mathbf{z})$, $\log p_\theta (\mathbf{z})$, and $\log q(\mathbf{z} | \mathbf{x})$ together, we arrive at

\textbf{Bernoulli Output Distribution}:
\begin{equation}
    \nabla_{\bm{\mu}_q} \mathcal{L} = \mathbb{E}_{\mathcal{N}(\bm{\epsilon}; \mathbf{0}, \mathbf{I})} \left[ {\frac{\partial \bm{\mu}_\mathbf{x}}{\partial \bm{\mu}_q}}^\intercal \frac{\mathbf{x} - \bm{\mu}_\mathbf{x}}{\bm{\mu}_\mathbf{x} \odot (\mathbf{1} - \bm{\mu}_\mathbf{x})} -\frac{\bm{\mu}_q + \bm{\sigma}_q \odot \bm{\epsilon} - \bm{\mu}_p}{\bm{\sigma}^2_p} \right]
\end{equation}
\begin{equation}
    \nabla_{\bm{\sigma}^2_q} \mathcal{L} = \mathbb{E}_{\mathcal{N}(\bm{\epsilon}; \mathbf{0}, \mathbf{I})} \left[ {\frac{\partial \bm{\mu}_\mathbf{x}}{\partial \bm{\sigma}^2_q}}^\intercal \frac{\mathbf{x} - \bm{\mu}_\mathbf{x}}{\bm{\mu}_\mathbf{x} \odot (\mathbf{1} - \bm{\mu}_\mathbf{x})} - \left( \diag{\frac{\bm{\epsilon}}{2 \bm{\sigma}_q}} \right)^\intercal \frac{\bm{\mu}_q + \bm{\sigma}_q \odot \bm{\epsilon} - \bm{\mu}_p}{\bm{\sigma}^2_p} \right] -\frac{\mathbf{1}}{2 \bm{\sigma}^2_q}
\end{equation}

\textbf{Gaussian Output Distribution}:
\begin{equation}
    \nabla_{\bm{\mu}_q} \mathcal{L} = \mathbb{E}_{\mathcal{N} (\bm{\epsilon}; \mathbf{0}, \mathbf{I})} \left[ {\frac{\partial \bm{\mu}_\mathbf{x} }{\partial \bm{\mu}_q}}^\intercal \frac{\mathbf{x} - \bm{\mu}_\mathbf{x}}{\bm{\sigma}_\mathbf{x}^2} -\frac{\bm{\mu}_q + \bm{\sigma}_q \odot \bm{\epsilon} - \bm{\mu}_p}{\bm{\sigma}^2_p} \right]
\end{equation}
\begin{equation}
    \nabla_{\bm{\sigma}^2_q} \mathcal{L} = \mathbb{E}_{\mathcal{N} (\bm{\epsilon}; \mathbf{0}, \mathbf{I})} \left[ {\frac{\partial \bm{\mu}_\mathbf{x}}{\partial \bm{\sigma}^2_q}}^\intercal \frac{\mathbf{x} - \bm{\mu}_\mathbf{x}}{\bm{\sigma}_\mathbf{x}^2} - \left( \diag{\frac{\bm{\epsilon}}{2 \bm{\sigma}_q}} \right)^\intercal \frac{\bm{\mu}_q + \bm{\sigma}_q \odot \bm{\epsilon} - \bm{\mu}_p}{\bm{\sigma}^2_p} \right] -\frac{\mathbf{1}}{2 \bm{\sigma}^2_q}
\end{equation}

\subsection{Approximate Posterior Gradients in Hierarchical Latent Variable Models}
\label{hierarchical gradients}

\begin{figure*}[t!]
    \centering
    \includegraphics[width=0.4\textwidth]{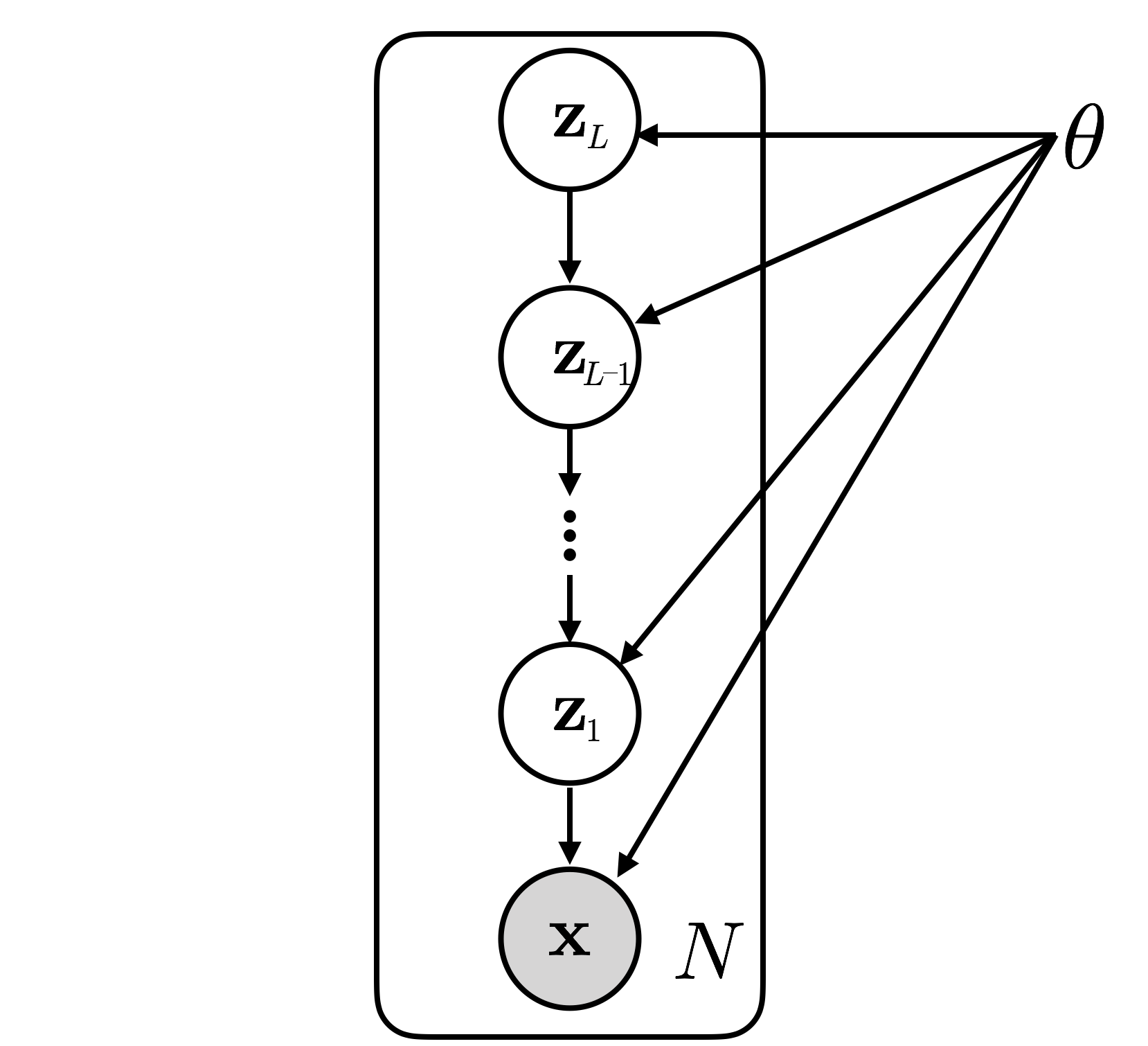}
    \caption{Plate notation for a hierarchical latent variable model consisting of $L$ levels of latent variables. Variables at higher levels provide empirical priors on variables at lower levels. With data-dependent priors, the model has more flexibility.}
    \label{fig: hierarchical latent variable model}
\end{figure*}

Hierarchical latent variable models factorize the latent variables over multiple levels, $\mathbf{z} = \{ \mathbf{z}_1, \mathbf{z}_2, \dots, \mathbf{z}_L\}$. Latent variables at higher levels provide \textit{empirical priors} on latent variables at lower levels. Here, we assume a first-order Markov graphical structure, as shown in Figure \ref{fig: hierarchical latent variable model}, though more general structures are possible. For an intermediate latent level, we use the notation $q(\mathbf{z}_\ell | \cdot) = \mathcal{N} (\mathbf{z}_\ell; \bm{\mu}_{\ell, q}, \diag{\bm{\sigma}^2_{\ell, q}})$ and $p(\mathbf{z}_\ell | \mathbf{z}_{\ell+1}) = \mathcal{N} (\mathbf{z}_\ell; \bm{\mu}_{\ell, p}, \diag{\bm{\sigma}^2_{\ell, p}})$ to denote the approximate posterior and prior respectively. Analogously to the case of a Gaussian output density in a one-level model, the approximate posterior gradients at an intermediate level $\ell$ are:
\begin{equation}
    \nabla_{\bm{\mu}_{q, \ell}} \mathcal{L} = \mathbb{E}_{\mathcal{N} (\bm{\epsilon}; \mathbf{0}, \mathbf{I})} \bigg[ {\frac{\partial \bm{\mu}_{\ell-1, p} }{\partial \bm{\mu}_{\ell, q}}}^\intercal \frac{\bm{\mu}_{\ell-1, q} + \bm{\sigma}_{\ell-1, q} \odot \bm{\epsilon}_{\ell-1} - \bm{\mu}_{\ell-1, p}}{\bm{\sigma}_{\ell-1, p}^2} -\frac{\bm{\mu}_{\ell, q} + \bm{\sigma}_{\ell, q} \odot \bm{\epsilon}_\ell - \bm{\mu}_{\ell, p}}{\bm{\sigma}^2_{\ell, p}} \bigg],
\end{equation}

\begin{equation}
    \nabla_{\bm{\sigma}^2_q} \mathcal{L} = \mathbb{E}_{\mathcal{N} (\bm{\epsilon}; \mathbf{0}, \mathbf{I})} \bigg[ {\frac{\partial \bm{\mu}_{\ell-1, p} }{\partial \bm{\sigma}^2_{\ell, q}}}^\intercal \frac{\bm{\mu}_{\ell-1, q} + \bm{\sigma}_{\ell-1, q} \odot \bm{\epsilon}_{\ell-1} - \bm{\mu}_{\ell-1, p}}{\bm{\sigma}_{\ell-1, p}^2} - \left( \diag{\frac{\bm{\epsilon}_\ell}{2 \bm{\sigma}_{\ell, q}}} \right)^\intercal \frac{\bm{\mu}_{\ell, q} + \bm{\sigma}_{\ell, q} \odot \bm{\epsilon}_\ell - \bm{\mu}_{\ell, p}}{\bm{\sigma}^2_{\ell, p}} \bigg] -\frac{\mathbf{1}}{2 \bm{\sigma}^2_{\ell, q}}.
\end{equation}

The first terms inside each expectation are ``bottom-up" gradients coming from reconstruction errors at the level below. The second terms inside the expectations are ``top-down" gradients coming from priors generated by the level above. The last term in the variance gradient acts to reduce the entropy of the approximate posterior.

\section{Implementing Iterative Inference Models}
\label{appendix: implementation details}

Here, we provide specific implementation details for these models. Code for reproducing the experiments will be released online.

\subsection{Input Form}
Approximate posterior gradients and errors experience distribution shift during inference and training. Using these terms as inputs to a neural network can slow down and prevent training. For experiments on MNIST, we found the $\log$ transformation method proposed by \cite{andrychowicz2016learning} to work reasonably well: replacing $\nabla_{\bm{\lambda}} \mathcal{L}$ with the concatenation of $ \left[ \alpha \log (|\nabla_{\bm{\lambda}} \mathcal{L}| + \epsilon), \text{sign} (\nabla_{\bm{\lambda}} \mathcal{L}) \right]$, where $\alpha$ is a scaling constant and $\epsilon$ is a small constant for numerical stability. We also encode the current estimates of $\bm{\mu}_q$ and $\log \bm{\sigma}^2_q$. For experiments on CIFAR-10, we instead used layer normalization \cite{ba2016layer} to normalize each input to the iterative inference model. This normalizes each input over the non-batch dimension.

\subsection{Output Form}
For the output of these models, we use a gated updating scheme, where approximate posterior parameters are updated according to
\begin{equation}
    \bm{\lambda}_{t+1} = \mathbf{g}_t \odot \bm{\lambda}_{t} + (\mathbf{1} - \mathbf{g}_t) \odot f_t (\nabla_{\bm{\lambda}} \mathcal{L}, \bm{\lambda}_{t}; \phi).
\end{equation}
Here, $\odot$ represents element-wise multiplication and $\mathbf{g}_t = g_t (\nabla_{\bm{\lambda}} \mathcal{L}, \bm{\lambda}_{t}; \phi) \in \left[ 0, 1 \right]$ is the gating function for $\bm{\lambda}$ at time $t$, which we combine with the iterative inference model $f_t$. We found that this yielded improved performance and stability over the additive updating scheme used in \cite{andrychowicz2016learning}.

\subsection{Training}

To train iterative inference models for latent Gaussian models, we use stochastic estimates of $\nabla_\phi \mathcal{L}$ from the reparameterization trick. We accumulate these gradient estimates during inference, then update both $\phi$ and $\theta$ jointly. We train using a fixed number of inference iterations.

\begin{algorithm}[tb]
   \caption{Iterative Amortized Inference}
   \label{alg:example}
\begin{algorithmic}
   \STATE {\bfseries Input:} data $\mathbf{x}$, generative model $p_\theta (\mathbf{x} , \mathbf{z})$, inference model $f$
   \STATE Initialize $t=0$ 
   \STATE Initialize $\nabla_\phi = 0$
   \STATE Initialize $q (\mathbf{z} | \mathbf{x})$ with $\bm{\lambda}_0$
   \REPEAT
   \STATE Sample $\mathbf{z} \sim q (\mathbf{z} | \mathbf{x})$
   \STATE Evaluate $\mathcal{L}_t = \mathcal{L} (\mathbf{x}, \bm{\lambda}_{t}; \theta)$
   \STATE Calculate $\nabla_{\bm{\lambda}} \mathcal{L}_t$ and $\nabla_{\phi} \mathcal{L}_t$
   % \STATE Calculate $\nabla_{\bm{\lambda}} \mathcal{L}_t$
   \STATE Update $\bm{\lambda}_{t+1} = f_t (\nabla_{\bm{\lambda}} \mathcal{L}_t, \bm{\lambda}_{t}; \phi)$
   \STATE $t = t + 1$
   \STATE $\nabla_\phi = \nabla_\phi + \nabla_{\phi} \mathcal{L}_t$
   \UNTIL{$\mathcal{L}$ converges}
   \STATE $\theta = \theta + \alpha_\theta \nabla_\theta \mathcal{L}$
   \STATE $\phi = \phi + \alpha_\phi \nabla_\phi$
\end{algorithmic}
\end{algorithm}

\section{Experiment Details}
\label{appendix: experiment details}

Inference model and generative model parameters ($\phi$ and $\theta$) were trained jointly using the adam optimizer \cite{kingma2014adam}. The learning rate was set to 0.0002 for both sets of parameters and all other optimizer parameters were set to their default values. Learning rates were decayed exponentially by a factor of 0.999 each epoch. All models utilized exponential linear unit (ELU) activation functions \cite{clevert2015fast}, although we found other non-linearities to work as well. Unless otherwise stated, all inference models were symmetric to their corresponding generative models. Iterative inference models for all experiments were implemented as feed-forward networks to make comparison with standard inference models easier.

\subsection{Two-Dimensional Latent Gaussian Models}
\label{appendix: 2d latent gaussian}

We trained models with 2 latent dimensions and a point estimate approximate posterior. That is, $q(\mathbf{z} | \mathbf{x}) = \delta (\mathbf{z} = \bm{\mu}_q)$ is a Dirac delta function at the point $\bm{\mu}_q = (\mu_1, \mu_2)$. We trained these models on binarized MNIST. The generative models consisted of a neural network with 2 hidden layers, each with 512 units. The output of the generative model was the mean of a Bernoulli distribution. The optimization surface of each model was evaluated on a grid of range [-5, 5] in increments of 0.05 for each latent variable. The iterative inference model shown in Figure 3 encodes $\mathbf{x}$, $\bm{\varepsilon}_\mathbf{x}$, and $\bm{\varepsilon}_\mathbf{z}$.

\subsection{$\mathcal{L}$ During Inference}
\label{appendix: ELBO during inference}

We trained one-level models on MNIST using iterative inference models that encode gradients ($\nabla_{\bm{\lambda}} \mathcal{L}$) for 16 iterations. We compared against stochastic gradient descent (SGD), SGD with momentum, RMSProp, and Adam, using learning rates in $\{0.5, 0.4, 0.3, 0.2, 0.1, 0.01, 0.001 \}$ and taking the best result. In addition to performance over iterations, we also compared the optimization techniques on the basis of wall clock time. Despite requiring more time per inference iteration, we observed that the iterative inference model still outperformed the conventional optimization techniques.

\subsection{Reconstructions Over Inference Iterations}
\label{appendix: reconstructions over inference iterations}

We trained iterative inference models on MNIST, Omniglot, and SVHN by encoding approximate posterior gradients ($\nabla_{\bm{\lambda}} \mathcal{L}$) for 16 iterations. For CIFAR-10, we trained an iterative inference model by encoding errors for 10 inference iterations. For MNIST and Omniglot, we used a generative model architecture with 2 hidden layers, each with 512 units, a latent space of size 64, and a symmetric iterative inference model. For SVHN and CIFAR-10, we used 3 hidden layers in the iterative inference and 1 in the generative model, with 2,048 units at each hidden layer and a latent space of size 1,024.

\begin{figure*}[h]
    \centering
    \begin{subfigure}[t]{0.6\textwidth}
        \centering
        \includegraphics[width=\textwidth]{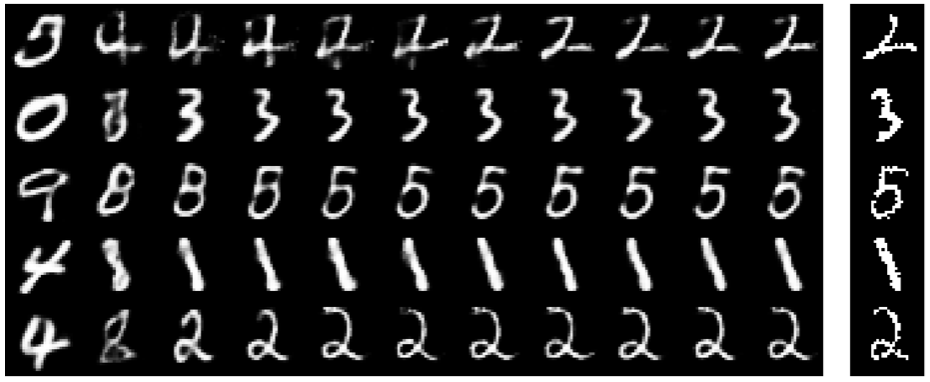}
        % \caption{}
    \end{subfigure}%
    
    \begin{subfigure}[t]{0.6\textwidth}
        \centering
        \includegraphics[width=\textwidth]{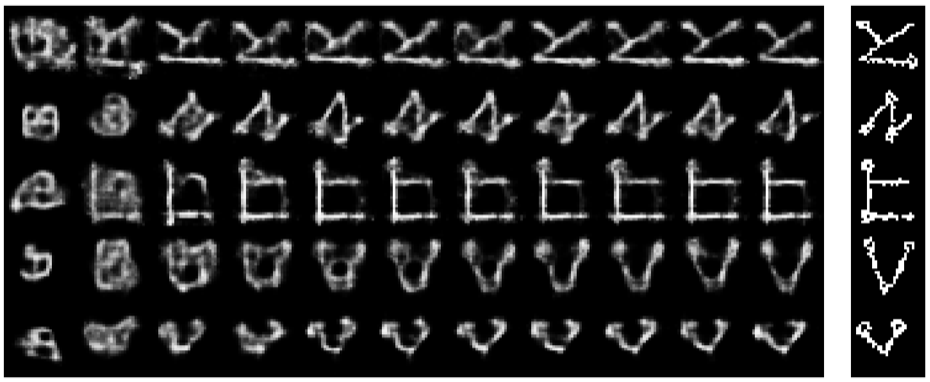}
        % \caption{}
    \end{subfigure}%
    
    \begin{subfigure}[t]{0.6\textwidth}
        \centering
        \includegraphics[width=\textwidth]{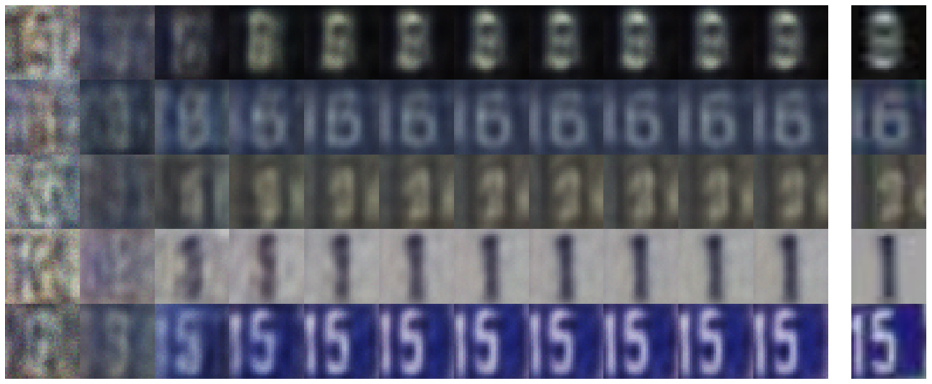}
        % \caption{}
    \end{subfigure}%
    
    \begin{subfigure}[t]{0.6\textwidth}
        \centering
        \includegraphics[width=\textwidth]{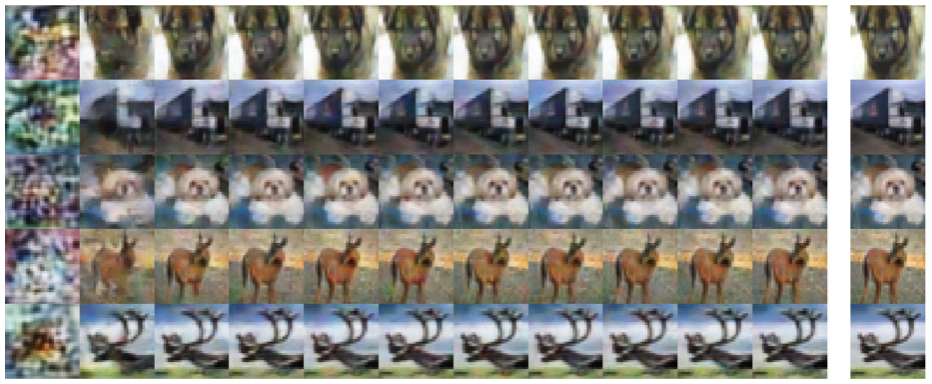}
        % \caption{}
    \end{subfigure}
    \caption{Reconstructions over inference iterations (left to right) for examples from (top to bottom) MNIST, Omniglot, SVHN, and CIFAR-10. Corresponding data examples are shown on the right of each panel.}
    \label{fig: additional reconstructions over iterations}
\end{figure*}

\subsection{Gradient Magnitudes}
\label{appendix: gradient magnitudes}

While training iterative inference models, we recorded approximate posterior gradient magnitudes at each inference iteration. We observed that, on average, the magnitudes decreased during inference optimization. This decrease was more prevalent for the approximate posterior mean gradients. For Figure 6, we trained an iterative inference model on RCV1 by encoding gradients ($\nabla_{\bm{\lambda}} \mathcal{L}$) for 16 inference iterations. The generative model contained a latent variable of size 512 and 2 fully-connected layers of 512 units each. The inference model was symmetric.

\subsection{Additional Inference Iterations}
\label{appendix: additional inference iterations}

We used an architecture of 2 hidden layers, each with 512 units, for the output model and inference models. The latent variable contained 64 dimensions. We trained all models for 1,500 epochs. We were unable to run multiple trials for each experimental set-up, but on a subset of runs for standard and iterative inference models, we observed that final performance had a standard deviation less than 0.1 nats, below the difference in performance between models trained with different numbers of inference iterations.

\subsection{Additional Latent Samples}
\label{appendix: additional latent samples}

We used an architecture of 2 hidden layers, each with 512 units, for the output model and inference models. The latent variable contained 64 dimensions. Each model was trained by drawing the corresponding number of samples from the approximate posterior distribution to obtain ELBO estimates and gradients. Iterative inference models were trained by encoding the data ($\mathbf{x}$) and the approximate posterior gradients ($\nabla_{\bm{\lambda}} \mathcal{L}$) for 5 inference iterations. All models were trained for 1,500 epochs.

\subsection{Comparison with Standard Inference Models}
\label{appendix: comparison with standard inference models}

\subsubsection{MNIST}

For MNIST, one-level models consisted of a latent variable of size 64, and the inference and generative networks both consisted of 2 hidden layers, each with 512 units. Hierarchical models consisted of 2 levels with latent variables of size 64 and 32 in hierarchically ascending order. At each level, the inference and generative networks consisted of 2 hidden layers, with 512 units at the first level and 256 units at the second level. At the first level of latent variables, we also used a set of deterministic units, also of size 64, in both the inference and generative networks. Hierarchical models included batch normalization layers at each hidden layer of the inference and generative networks; we found this beneficial for training both standard and iterative inference models. Both encoder and decoder networks in the hierarchical model utilized highway skip connections at each layer at both levels. Iterative models were trained by encoding data and errors for 5 inference iterations.

\subsubsection{CIFAR-10}

For CIFAR-10, one-level models consisted of a latent variable of size 1,024, an encoder network with 3 hidden layers of 2,048 units, and a decoder network with 1 hidden layer with 2,048 units. We found this set-up performed better than a symmetric encoder and decoder for both standard and iterative inference models. Hierarchical models were the same as the one-level model, adding another latent variable of size 512, with another 3 layer encoder of with 1,024 units and a 1 layer decoder with 1,024 units. Both encoder and decoder networks in the hierarchical model utilized highway skip connections at each layer at both levels. Models were all trained for 150 epochs. We annealed the KL-divergence term during the first 50 epochs when training hierarchical models. Iterative inference models were trained by encoding the data and gradients for 5 inference iterations.

\subsubsection{RCV1}

We followed the same processing procedure as \cite{krishnan2017challenges}, encoding data using normalized TF-IDF features. For encoder and decoder, we use 2-layer networks, each with 2,048 units and ELU non-linearities. We use a latent variable of size 1,024. The iterative inference model was trained by encoding gradients for 10 steps. Both models were trained using 5 approximate posterior samples at each iteration. We evaluate the models by reporting perplexity on the test set (Table 2). Perplexity, $P$, is defined as
\begin{equation}
    P \equiv \exp (-\frac{1}{N} \sum_i \frac{1}{N_i} \log p (\mathbf{x}^{(i)})),
\end{equation}
where $N$ is the number of examples and $N_i$ is the total number of word counts in example $i$. We evaluate perplexity by estimating each $\log p ( \mathbf{x}^{(i)} )$ with 5,000 importance weighted samples. We also report an upper bound on perplexity using $\mathcal{L}$.

\bibliography{appendix_bib}

\begin{thebibliography}{30}
\providecommand{\natexlab}[1]{#1}
\providecommand{\url}[1]{\texttt{#1}}
\expandafter\ifx\csname urlstyle\endcsname\relax
  \providecommand{\doi}[1]{doi: #1}\else
  \providecommand{\doi}{doi: \begingroup \urlstyle{rm}\Url}\fi

\bibitem[Andrychowicz et~al.(2016)Andrychowicz, Denil, Gomez, Hoffman, Pfau,
  Schaul, and de~Freitas]{andrychowicz2016learning}
Andrychowicz, M., Denil, M., Gomez, S., Hoffman, M.~W., Pfau, D., Schaul, T.,
  and de~Freitas, N.
\newblock Learning to learn by gradient descent by gradient descent.
\newblock In \emph{Advances in Neural Information Processing Systems (NIPS)},
  pp.\  3981--3989, 2016.

\bibitem[Ba et~al.(2016)Ba, Kiros, and Hinton]{ba2016layer}
Ba, J.~L., Kiros, J.~R., and Hinton, G.~E.
\newblock Layer normalization.
\newblock \emph{arXiv preprint arXiv:1607.06450}, 2016.

\bibitem[Clevert et~al.(2015)Clevert, Unterthiner, and
  Hochreiter]{clevert2015fast}
Clevert, D.-A., Unterthiner, T., and Hochreiter, S.
\newblock Fast and accurate deep network learning by exponential linear units
  (elus).
\newblock \emph{arXiv preprint arXiv:1511.07289}, 2015.

\bibitem[Cremer et~al.(2017)Cremer, Li, and Duvenaud]{cremerinference}
Cremer, C., Li, X., and Duvenaud, D.
\newblock Inference suboptimality in variational autoencoders.
\newblock \emph{NIPS Workshop on Advances in Approximate Bayesian Inference},
  2017.

\bibitem[Dayan et~al.(1995)Dayan, Hinton, Neal, and Zemel]{dayan1995helmholtz}
Dayan, P., Hinton, G.~E., Neal, R.~M., and Zemel, R.~S.
\newblock The helmholtz machine.
\newblock \emph{Neural computation}, 7\penalty0 (5):\penalty0 889--904, 1995.

\bibitem[Dempster et~al.(1977)Dempster, Laird, and Rubin]{dempster1977maximum}
Dempster, A.~P., Laird, N.~M., and Rubin, D.~B.
\newblock Maximum likelihood from incomplete data via the em algorithm.
\newblock \emph{Journal of the royal statistical society. Series B
  (methodological)}, pp.\  1--38, 1977.

\bibitem[Gershman \& Goodman(2014)Gershman and Goodman]{gershman2014amortized}
Gershman, S. and Goodman, N.
\newblock Amortized inference in probabilistic reasoning.
\newblock In \emph{Proceedings of the Cognitive Science Society}, volume~36,
  2014.

\bibitem[Gregor et~al.(2014)Gregor, Danihelka, Mnih, Blundell, and
  Wierstra]{gregor2013deep}
Gregor, K., Danihelka, I., Mnih, A., Blundell, C., and Wierstra, D.
\newblock Deep autoregressive networks.
\newblock In \emph{Proceedings of the International Conference on Machine
  Learning (ICML)}, pp.\  1242--1250, 2014.

\bibitem[Gregor et~al.(2015)Gregor, Danihelka, Graves, Rezende, and
  Wierstra]{gregor2015draw}
Gregor, K., Danihelka, I., Graves, A., Rezende, D.~J., and Wierstra, D.
\newblock Draw: A recurrent neural network for image generation.
\newblock \emph{Proceedings of the International Conference on Machine Learning
  (ICML)}, pp.\  1462--1471, 2015.

\bibitem[Gregor et~al.(2016)Gregor, Besse, Rezende, Danihelka, and
  Wierstra]{gregor2016towards}
Gregor, K., Besse, F., Rezende, D.~J., Danihelka, I., and Wierstra, D.
\newblock Towards conceptual compression.
\newblock In \emph{Advances In Neural Information Processing Systems (NIPS)},
  pp.\  3549--3557, 2016.

\bibitem[Higgins et~al.(2016)Higgins, Matthey, Pal, Burgess, Glorot, Botvinick,
  Mohamed, and Lerchner]{higgins2016beta}
Higgins, I., Matthey, L., Pal, A., Burgess, C., Glorot, X., Botvinick, M.,
  Mohamed, S., and Lerchner, A.
\newblock beta-vae: Learning basic visual concepts with a constrained
  variational framework.
\newblock In \emph{Proceedings of the International Conference on Learning
  Representations (ICLR)}, 2016.

\bibitem[Hjelm et~al.(2016)Hjelm, Salakhutdinov, Cho, Jojic, Calhoun, and
  Chung]{hjelm2016iterative}
Hjelm, D., Salakhutdinov, R.~R., Cho, K., Jojic, N., Calhoun, V., and Chung, J.
\newblock Iterative refinement of the approximate posterior for directed belief
  networks.
\newblock In \emph{Advances in Neural Information Processing Systems (NIPS)},
  pp.\  4691--4699, 2016.

\bibitem[Hoffman et~al.(2013)Hoffman, Blei, Wang, and
  Paisley]{hoffman2013stochastic}
Hoffman, M.~D., Blei, D.~M., Wang, C., and Paisley, J.
\newblock Stochastic variational inference.
\newblock \emph{The Journal of Machine Learning Research}, 14\penalty0
  (1):\penalty0 1303--1347, 2013.

\bibitem[Jordan et~al.(1998)Jordan, Ghahramani, Jaakkola, and
  Saul]{jordan1998introduction}
Jordan, M.~I., Ghahramani, Z., Jaakkola, T.~S., and Saul, L.~K.
\newblock An introduction to variational methods for graphical models.
\newblock \emph{NATO ASI SERIES D BEHAVIOURAL AND SOCIAL SCIENCES},
  89:\penalty0 105--162, 1998.

\bibitem[Karl et~al.(2017)Karl, Soelch, Bayer, and van~der Smagt]{karl2016deep}
Karl, M., Soelch, M., Bayer, J., and van~der Smagt, P.
\newblock Deep variational bayes filters: Unsupervised learning of state space
  models from raw data.
\newblock In \emph{Proceedings of the International Conference on Learning
  Representations (ICLR)}, 2017.

\bibitem[Kim et~al.(2018)Kim, Wiseman, Miller, Sontag, and Rush]{kim2018semi}
Kim, Y., Wiseman, S., Miller, A.~C., Sontag, D., and Rush, A.~M.
\newblock Semi-amortized variational autoencoders.
\newblock In \emph{Proceedings of the International Conference on Machine
  Learning (ICML)}, 2018.

\bibitem[Kingma \& Ba(2014)Kingma and Ba]{kingma2014adam}
Kingma, D. and Ba, J.
\newblock Adam: A method for stochastic optimization.
\newblock \emph{arXiv preprint arXiv:1412.6980}, 2014.

\bibitem[Kingma \& Welling(2014)Kingma and Welling]{kingma2014stochastic}
Kingma, D.~P. and Welling, M.
\newblock Stochastic gradient vb and the variational auto-encoder.
\newblock In \emph{Proceedings of the International Conference on Learning
  Representations (ICLR)}, 2014.

\bibitem[Krishnan et~al.(2018)Krishnan, Liang, and
  Hoffman]{krishnan2017challenges}
Krishnan, R.~G., Liang, D., and Hoffman, M.
\newblock On the challenges of learning with inference networks on sparse,
  high-dimensional data.
\newblock In \emph{Proceedings of the International Conference on Artificial
  Intelligence and Statistics (AISTATS)}, pp.\  143--151, 2018.

\bibitem[Krizhevsky \& Hinton(2009)Krizhevsky and
  Hinton]{krizhevsky2009learning}
Krizhevsky, A. and Hinton, G.
\newblock Learning multiple layers of features from tiny images.
\newblock 2009.

\bibitem[Lake et~al.(2013)Lake, Salakhutdinov, and Tenenbaum]{lake2013one}
Lake, B.~M., Salakhutdinov, R.~R., and Tenenbaum, J.
\newblock One-shot learning by inverting a compositional causal process.
\newblock In \emph{Advances in Neural Information Processing Systems (NIPS)},
  pp.\  2526--2534, 2013.

\bibitem[LeCun et~al.(1998)LeCun, Bottou, Bengio, and
  Haffner]{lecun1998gradient}
LeCun, Y., Bottou, L., Bengio, Y., and Haffner, P.
\newblock Gradient-based learning applied to document recognition.
\newblock \emph{Proceedings of the IEEE}, 86\penalty0 (11):\penalty0
  2278--2324, 1998.

\bibitem[Lewis et~al.(2004)Lewis, Yang, Rose, and Li]{lewis2004rcv1}
Lewis, D.~D., Yang, Y., Rose, T.~G., and Li, F.
\newblock Rcv1: A new benchmark collection for text categorization research.
\newblock \emph{The Journal of Machine Learning Research}, 5\penalty0
  (Apr):\penalty0 361--397, 2004.

\bibitem[Neal \& Hinton(1998)Neal and Hinton]{neal1998view}
Neal, R.~M. and Hinton, G.~E.
\newblock A view of the em algorithm that justifies incremental, sparse, and
  other variants.
\newblock In \emph{Learning in graphical models}, pp.\  355--368. Springer,
  1998.

\bibitem[Netzer et~al.(2011)Netzer, Wang, Coates, Bissacco, Wu, and
  Ng]{netzer2011reading}
Netzer, Y., Wang, T., Coates, A., Bissacco, A., Wu, B., and Ng, A.~Y.
\newblock Reading digits in natural images with unsupervised feature learning.
\newblock In \emph{NIPS workshop on deep learning and unsupervised feature
  learning}, 2011.

\bibitem[Putzky \& Welling(2017)Putzky and Welling]{putzky2017recurrent}
Putzky, P. and Welling, M.
\newblock Recurrent inference machines for solving inverse problems.
\newblock \emph{arXiv preprint arXiv:1706.04008}, 2017.

\bibitem[Ranganath et~al.(2014)Ranganath, Gerrish, and
  Blei]{ranganath2014black}
Ranganath, R., Gerrish, S., and Blei, D.
\newblock Black box variational inference.
\newblock In \emph{Proceedings of the International Conference on Artificial
  Intelligence and Statistics (AISTATS)}, pp.\  814--822, 2014.

\bibitem[Rezende et~al.(2014)Rezende, Mohamed, and
  Wierstra]{rezende2014stochastic}
Rezende, D.~J., Mohamed, S., and Wierstra, D.
\newblock Stochastic backpropagation and approximate inference in deep
  generative models.
\newblock In \emph{Proceedings of the International Conference on Machine
  Learning (ICML)}, pp.\  1278--1286, 2014.

\bibitem[S{\o}nderby et~al.(2016)S{\o}nderby, Raiko, Maal{\o}e, S{\o}nderby,
  and Winther]{sonderby2016ladder}
S{\o}nderby, C.~K., Raiko, T., Maal{\o}e, L., S{\o}nderby, S.~K., and Winther,
  O.
\newblock Ladder variational autoencoders.
\newblock In \emph{Advances in Neural Information Processing Systems (NIPS)},
  pp.\  3738--3746, 2016.

\bibitem[Xue et~al.(2016)Xue, Wu, Bouman, and Freeman]{xue2016visual}
Xue, T., Wu, J., Bouman, K., and Freeman, B.
\newblock Visual dynamics: Probabilistic future frame synthesis via cross
  convolutional networks.
\newblock In \emph{Advances in Neural Information Processing Systems (NIPS)},
  pp.\  91--99, 2016.

\end{thebibliography}
\bibliographystyle{icml2018}

\end{document}